\documentclass[final,3p,times]{elsarticle}

\usepackage{amssymb}
\usepackage{amsmath}
\usepackage{placeins}
\usepackage{multirow}
\usepackage{booktabs}
\usepackage{enumitem}
\usepackage[hidelinks]{hyperref}

\begin{document}

\begin{frontmatter}



\title{Availability-Aware Historical Feedback Learning for Electric Arc Furnace Temperature and Chemical Composition Prediction}


\author[inst1]{Ahmed Shoyeb Raihan}
\author[inst1]{Farzana Islam}
\author[inst1]{Ashish Nimbarte}
\author[inst1]{Imtiaz Ahmed\corref{cor1}}

\ead{imtiaz.ahmed@mail.wvu.edu}

\cortext[cor1]{Corresponding author}

\affiliation[inst1]{
    organization={Department of Industrial and Management Systems Engineering, West Virginia University},
    city={Morgantown},
    postcode={26506},
    state={West Virginia},
    country={USA}
}




\begin{abstract}

Electric arc furnace (EAF) steelmaking plays an increasingly important role in recycled and lower-carbon steel production. However, variability in charged materials, energy input, and operating conditions makes accurate prediction of molten steel temperature and final chemical composition challenging. These critical process outcomes influence subsequent refining operations and steel quality, yet their measurements are often intermittent or delayed, limiting the information available for timely prediction. Although machine learning (ML) approaches have demonstrated considerable potential for EAF process modeling, the systematic incorporation of previously measured outcomes according to their availability at the prediction time remains an important research opportunity. To address this challenge, this study develops an availability-aware historical-feedback learning framework for predicting EAF endpoint temperature and the final concentrations of twelve chemical elements. Eight industrial data sources are integrated into three progressively enriched predictor representations comprising 19 static, 39 temporal, and 47 history-informed features. Historical outcomes are incorporated only when eligible under an assumed 60-minute measurement-release delay. Using 19,738 temperature heats and 3,341 chemistry heats, twelve tuned regression algorithms are evaluated through chronological forward validation. The proposed historical-feedback representation reduces root mean squared error (RMSE) in 147 of 156 matched algorithm--target comparisons. For temperature prediction, ExtraTrees achieves an RMSE of 12.290~$^\circ$C, representing a 5.54\% reduction compared with its temporal-only counterpart. Historical-feature ablation and feedback-delay sensitivity analyses further demonstrate the importance of recent outcome information and its timely availability. In a separate retrospective later-period evaluation, lightweight adaptive residual correction reduces temperature RMSE from 12.502 to 12.275~$^\circ$C without retraining the underlying model. Exploratory deep learning (DL) comparisons and target-specific feature-attribution analyses provide additional insights into predictive behavior. Overall, the findings establish historical outcome availability and sequential feedback as important considerations for EAF endpoint prediction, providing a unified framework for developing more informative and adaptive data-driven steelmaking models.

\end{abstract}




\begin{keyword}
Electric Arc Furnace \sep
Tapping Temperature Prediction \sep
Steel Chemical Composition \sep
Historical Outcome Feedback \sep
Availability-Aware Machine Learning \sep
Adaptive Residual Correction




\end{keyword}

\end{frontmatter}




\section{Introduction}
\label{sec1}

Electric arc furnace (EAF) steelmaking is an increasingly important production route for recycled and lower-carbon steel \cite{fan2021low,lee2014review}. Its industrial operation involves rapid changes in electrical input, material charging, gas and carbon injection, and melt chemistry, while critical metallurgical outcomes are measured intermittently rather than continuously \cite{choi2023machine,shyamal2019real,manojlovic2022machine}. Among these outcomes, tapping temperature and final elemental composition are particularly consequential: they characterize the thermal and chemical state of the steel delivered to subsequent refining and casting stages. Reliable advance estimates can provide decision-relevant information when direct outcome measurements are sparse, delayed, or obtained close to the end of a heat. Industrial reviews of furnace modeling and instrumentation further document the diverse measurement technologies and process information needed to observe EAF operation \cite{abadi2024review,rautioaho2024electric}. The scientific challenge is therefore not only to fit accurate predictive relationships, but also to construct those relationships from information that is legitimately available \emph{at the prediction instant}.

EAF endpoint prediction has benefited from advances in industrial machine learning (ML). Studies have demonstrated the potential of engineered process descriptors, nonlinear regression, and model interpretation for tapping-temperature estimation \cite{choi2023machine,song2023data,yang2025eaf_datamining,lu2026eaf_interpretable}. Parallel work has developed data-driven and hybrid predictors for final carbon, phosphorus, and other steel-quality variables \cite{yang2022influence,zou2022prediction,wei2018hybrid,azzaz2025prediction}. Sequence-based models also offer ways to learn temporal patterns in furnace operation \cite{li2021long,klimas2023application,raihan2023bilstm, raihan2026hybrid}. Taken together, these developments establish that substantial predictive information is present in EAF records. They also motivate a more specific question: \emph{how much additional information can be recovered from the measured outcomes of preceding heats, and how does that contribution depend on when those outcomes become available?}

This question differs from conventional within-heat time-series modeling. Operational readings collected during the current heat capture its evolving process conditions, whereas the measured temperature or chemistry of a preceding heat becomes a usable feedback signal only after the corresponding result is released. Treating all previous heat labels as immediately accessible can produce a misleading representation of the information available for a prospective prediction. Conversely, discarding earlier outcomes altogether neglects a potentially informative record of recent process behavior. Work on learning from data streams and concept drift establishes the broader importance of delayed labels, sequential updates, and changing data-generating conditions \cite{cacciarelli2024active_streams,gama2014concept_drift}. In EAF steelmaking, these concepts suggest a prediction formulation in which \emph{measurement availability itself} is part of the feature-engineering and evaluation protocol.

We investigate this formulation through a unified, heat-indexed predictive framework for tapping temperature and twelve final chemical elements: carbon, silicon, manganese, phosphorus, sulfur, copper, chromium, molybdenum, nickel, arsenic, tin, and nitrogen. Eight categories of historical industrial records are integrated into task-specific modeling cohorts, with operational predictors assembled relative to a defined prediction cutoff. Three nested information representations are then compared: \textsc{Static}, containing 19 pre-cutoff operational descriptors; \textsc{Temporal}, adding 20 within-heat summaries; and \textsc{History}, further adding eight descriptors of earlier outcomes eligible under an assumed measurement-release rule. These historical descriptors summarize recent results, short- and longer-term averages, variability, and the freshness of available feedback. The representation is target-specific but model-independent, allowing its effect to be evaluated without attributing a feature-engineering benefit to the selection of a particular learning algorithm.

The experimental design is constructed to distinguish these sources of predictive information. Twelve regression families are evaluated under three expanding-window chronological validation folds with hyperparameter tuning performed within each outer training period. Comparisons between \textsc{Temporal} and \textsc{History} are paired for the same algorithm, target, and validation heats, while simple historical predictors establish whether the added value extends beyond repeating recently observed outcomes. Across the completed development-period experiments, the \textsc{History} representation yielded lower root mean squared error (RMSE) in 147 of 156 matched algorithm--target comparisons. This breadth of improvement points to the importance of accessible historical outcomes as an information source across model families, rather than to the isolated superiority of one estimator. A complementary comparison with four neural network architectures examines whether more complex heat-sequence models provide an advantage on the available heat-level representations.

Beyond the principal benchmark, the proposed study examines \emph{why}, \emph{when}, and \emph{how} historical feedback contributes to prediction. Structured feature-group ablations distinguish the influence of the latest released result, moving historical averages, longer-window variation, and freshness-related descriptors. Outcome-release sensitivity experiments assess how the available history and resulting errors change when feedback is assumed to arrive after longer intervals. Finally, a lightweight adaptive extension uses released errors from earlier predictions to adjust later estimates while leaving the fitted base regressor unchanged. The latter mechanism is evaluated separately on a later chronological production period, together with diagnostics of large temperature prediction errors and target-specific feature attribution. This integrated treatment connects historical information design, time-aware assessment, and sequential correction in a single EAF prediction study.

The main contributions of this work are as follows:
\begin{enumerate}[label=\textbf{C\arabic*.},leftmargin=*,itemsep=4pt]

    \item \textbf{Cutoff-aware industrial data engineering and unified EAF endpoint modeling.} We develop a reproducible data-engineering pipeline for eight heterogeneous industrial sources, incorporating timestamp alignment, prediction-cutoff filtering, quality screening, and process-specific feature engineering. Target-specific data preparation and element-wise assay retention establish consistent modeling datasets for tapping temperature and twelve chemical outcomes, ensuring that predictions rely on information available at the defined prediction time.

    \item \textbf{Availability-aware historical-feedback representation.} We formulate an eight-descriptor outcome-history representation that incorporates only earlier measurements considered released by the current prediction cutoff. Combined with 19 static and 20 within-heat temporal predictors, it separates current-heat information from previously observed process outcomes and makes historical-feedback availability an explicit modeling dimension.

    \item \textbf{Systematic, algorithm-independent evaluation of historical information.} We compare twelve tuned regressors across three nested feature representations under chronological validation, with matched predictions, historical-reference baselines, and exploratory paired uncertainty estimates. This design quantifies the predictive contribution of released outcome information for temperature and all twelve chemical targets, beyond differences associated with model family alone.

    \item \textbf{Historical-feature contributions and feedback-timing analysis.} We examine the incremental contributions of historical feature groups, the consequences of delayed outcome release, chronological error behavior, and target-specific interpretability. These analyses reveal how predictive information depends on history composition and feedback freshness, rather than relying exclusively on aggregate model rankings.

    \item \textbf{Low-overhead sequential adaptation and broader architectural comparison.} We investigate an online correction based on previously released prediction errors, retaining a frozen base model while updating its feedback memory. Retrospective later-period evaluation characterizes the resulting improvements and target-dependent responses; an exploratory four-model deep learning (DL) comparison further contextualizes the classical history-informed predictors.
\end{enumerate}

Together, these contributions establish an integrated availability-aware learning methodology for EAF endpoint prediction, spanning raw industrial event processing, explicit prediction cutoffs, target-specific historical-outcome representations, and sequential inference. By separating within-heat measurements from previously released endpoint results and evaluating these information levels across twelve tuned regressors under chronological validation, the study demonstrates that the value of historical feedback extends across algorithms and chemical targets. Historical-feature ablation, release-delay sensitivity, and adaptive residual correction further explain how the available feedback contributes to predictive accuracy and how it can be used as production records evolve. The resulting framework connects industrial data engineering with systematic predictive learning for both thermal and chemical steel-quality outcomes.

The remainder of the paper is organized as follows. Section~\ref{sec:eaf_related_work} reviews related EAF prediction research and identifies the information-availability gap. Section~\ref{sec:eaf_data} describes the industrial records, task-specific prediction cutoffs, and modeling cohorts. Section~\ref{sec:eaf_framework} develops the historical-feedback representation and adaptive correction. Section~\ref{sec:eaf_experiments} presents the chronological evaluation and tuning protocols. Section~\ref{sec:results} reports the benchmark, ablation, robustness, adaptation, DL, and interpretability results. Section~\ref{sec:eaf_conclusion} presents the main conclusions, the implications for availability-aware EAF decision support, and directions for further research.

\section{Related Work and Research Gap}
\label{sec:eaf_related_work}

Data-driven modeling of EAF operations has progressed from individual process-variable estimators toward models integrating operational records, metallurgical knowledge, and explainable ML. The resulting literature establishes the value of predictive analytics for energy demand, thermal conditions, and steel quality. For the present study, the most relevant questions concern not only which regression algorithm is used, but also how information is assembled at the time of prediction, how successive heats provide new observations, and how the resulting predictions are evaluated chronologically. We organize the literature around these connections and use them to motivate the availability-aware historical-feedback formulation developed in Section~\ref{sec:eaf_framework}.

\subsection{Data-driven prediction of EAF temperature}
\label{subsec:eaf_related_temperature}

Tapping-temperature estimation is a well-established application of EAF process analytics because the thermal endpoint links melting and refining operations with subsequent steel handling. Earlier work has evaluated conventional ML predictors for tapping-temperature estimation and its relationship to electrical and other process inputs \cite{choi2023machine}. Song et al.~\cite{song2023data} developed an industrial-data-driven endpoint-temperature framework incorporating preprocessing, feature engineering, regression modeling, and sensitivity analysis. Such studies establish the practical value of combining operational measurements with data-driven models rather than considering temperature prediction as a purely algorithm-selection exercise. Early EAF temperature studies also investigated fuzzy-neural estimators, while industrial analyses have modeled furnace energy demand under incomplete knowledge of process parameters \cite{fernandez2008online,zawodnik2024tackling}. These parallel modeling tasks underscore the value of selecting information appropriate to the prediction objective \cite{ahmed2024spatiotemporal}.

Research has also explored sequential and hybrid representations of furnace behavior. Long short-term memory (LSTM) models have been investigated for EAF tapping-temperature prediction \cite{li2021long}, while attention-based recurrent networks and multivariate sequence models have been studied for furnace-wall or lining-temperature monitoring \cite{godoy2022attention,leon2021temperature}. These are related but distinct thermal prediction problems: wall-temperature trajectories and heat-level endpoint measurements have different targets, sensing conditions, and time scales. Other investigations have integrated mechanistic context, process indicators, and nonlinear prediction schemes to model EAF endpoints \cite{shi2025prediction,niyayesh2025predicting}. Related time-dependent studies have modeled EAF lining temperatures and heating-stage thermal behavior \cite{leon2020deep,li2022generative}. Temperature estimation has also been investigated during ladle treatment and molten-steel transfer \cite{wang2022dynamic,botnikov2019development,fang2023artificial}, illustrating the range of thermal targets encountered along the steelmaking route. These target definitions are distinct from the heat-level EAF endpoint considered here.

Recent studies further illustrate the continued development of feature-informed and interpretable EAF temperature models. Yang et al.~\cite{yang2025eaf_datamining} combined anomaly detection, feature selection, and supervised learning using plant data; Lu et al.~\cite{lu2026eaf_interpretable} examined a process-variable representation informed by metallurgical energy balance and model interpretation. These contributions underscore that high-performing endpoint models can be developed through several complementary routes, including feature engineering, nonlinear estimation, and metallurgical analysis. Our emphasis is different: we assess the incremental value of \emph{previously released endpoint outcomes}, holding the regression family and chronological evaluation conditions fixed when comparing feature representations. Consequently, the present results are intended as matched comparisons within one heat-indexed dataset, rather than numerical rankings against studies conducted with different furnaces or endpoint definitions.

\subsection{Prediction of final steel composition}
\label{subsec:eaf_related_chemistry}

Predicting the composition of tapped steel is similarly important because endpoint carbon, phosphorus, sulfur, alloying elements, and residual elements provide different forms of information about the resulting melt. Existing EAF studies have used artificial neural networks and hybrid process-informed methods to estimate endpoint carbon and phosphorus \cite{yang2022influence,zou2022prediction,wei2018hybrid}. Other research has investigated Bayesian-optimized and deeper neural estimators, as well as data-quality and optimization strategies, for carbon or phosphorus prediction \cite{zhu2024enhanced,lu2024catboost,azzaz2025prediction}. These methods demonstrate the usefulness of both process indicators and nonlinear learning for composition forecasting. Adjacent metallurgical prediction studies have modeled inclusion content in EAF steel, neural surrogates for slag chemical activities, and endpoint phosphorus in basic oxygen furnace operations \cite{ruiz2021machine,reinicke2024application,zhang2023comparison}. Although these applications concern different response variables or steelmaking stages, they illustrate the breadth of data-driven chemical-quality modeling.

The prediction setting becomes more intricate when the response comprises multiple chemical species. Different elemental targets may be characterized by different concentration ranges, assay frequencies, material histories, and predictive relationships. Research on tramp-element prediction using scrap-composition information \cite{schafer2025machine}, together with studies of multi-stage refining and carbon prediction \cite{hamachine,liu2024research}, highlights the importance of matching the feature representation to the chemistry outcome being estimated. Accordingly, our formulation treats the twelve final elemental measurements as \emph{twelve independently fitted target-specific regressions} within a common data and evaluation framework. This choice allows the same availability rules to be studied across elements while retaining observations for individual elements when another assay is missing; it does not claim that independent regressors explicitly represent cross-element dependencies.

Much of the composition literature focuses on predicting an individual element or on optimizing a particular predictive architecture. A complementary question is whether a single, clearly defined historical-information mechanism produces comparable benefits across a broader collection of chemical targets. The present study addresses this question through matched \textsc{Temporal} and \textsc{History} comparisons for each of the twelve elements, alongside temperature prediction, while reporting each outcome in its own physical units.

\subsection{Within-heat dynamics and previously released outcome feedback}
\label{subsec:eaf_related_feedback}

The temporal structure of EAF data has at least two distinct dimensions. \emph{Within-heat dynamics} describe changes in operational measurements, such as transformer activity and injection flows, during the current production heat. \emph{Across-heat feedback} instead describes the results of earlier heats that have become available before a new prediction is issued. Recurrent and sequence-aware EAF models have principally demonstrated the usefulness of time-dependent process measurements and learned temporal patterns \cite{li2021long,klimas2023application,son2021slag}. Summaries of short time windows and within-heat process stages provide a complementary representation for heat-indexed, tabular predictors \cite{al2024feature,manojlovic2022machine}. A related body of EAF analytics models energy consumption using deep networks and process-based features, including studies of electric-furnace energy demand and hybrid CONARC furnace operation \cite{chen2018energy,chavosh2022data,kota2022prediction}. Work on industrial energy-performance indicators and data-driven operating-period optimization provides additional examples of information extracted from heterogeneous process records \cite{khalid2024predictive,tomavzivc2022data}. These energy-oriented studies provide complementary process-data context rather than direct benchmark values for endpoint temperature or elemental composition.

For across-heat feedback, an additional distinction is critical: an earlier outcome can be used only after it is available, not simply because its heat has a smaller index. This is particularly relevant when laboratory measurements, quality inspections, or physical measurements are recorded with a delay. Work on learning from data streams explicitly recognizes the role of label-availability latency and its interaction with sequential learning \cite{cacciarelli2024active_streams}. A broader literature on concept drift and adaptive learning shows how a model's relationship to a changing process may evolve over time \cite{gama2014concept_drift}. These concepts motivate a representation that tracks both the \emph{values} of prior outcomes and the \emph{freshness} of released information.

The proposed framework augments current-heat operational and temporal features with eight descriptors computed from previously released values of the \emph{same prediction target}. These include recent values, short- and long-window averages, historical variation, and indicators of outcome freshness and availability. The modeling question is whether this explicitly release-aware representation adds predictive value over operational and within-heat temporal features alone. The assumed measurement-release delay is evaluated through sensitivity analyses rather than treated as an observed plant reporting interval.

\subsection{Chronological evaluation, adaptive prediction, and interpretability}
\label{subsec:eaf_related_evaluation}

The evaluation protocol plays an essential role when consecutive heats are temporally ordered and the information set changes over time. Cerqueira et al.~\cite{cerqueira2020time_series_evaluation} showed that different forecasting evaluation schemes can yield materially different performance estimates, especially for nonstationary sequences. A forward-validation protocol therefore provides an appropriate basis for asking whether a historical feature available at one cutoff continues to be useful for later heats. In our study, chronological outer folds assess each prespecified algorithm, with hyperparameter selection performed only on earlier training partitions. Comparisons between feature configurations are matched by target, algorithm, and validation heat.

Beyond fixed prediction models, the ability to observe new outcomes suggests lightweight adaptation. Concept-drift research distinguishes updating a model itself from adjusting predictions using information arriving through a stream \cite{gama2014concept_drift}. We investigate the latter mechanism by maintaining a frozen history-informed regressor and applying a calibrated additive residual correction derived from previously released \emph{base-model residuals}. The correction is updated sequentially without retraining the regression model, and its contribution is assessed separately from the eight historical-outcome descriptors. This distinguishes model updating from feedback-based adjustment while retaining a transparent, testable inference sequence.

Interpretability supplies a complementary perspective on how the model uses operational and historical inputs. In EAF applications, feature engineering and Shapley additive explanations (SHAP) have been applied to energy-demand and operating-variable models \cite{carlsson2020interpretable,manojlovic2022machine}. SHAP provides a general additive feature-attribution framework \cite{lundberg2017shap}; it describes associations within a fitted predictive model rather than the causal effects of changing furnace settings. The interpretation of Shapley-based explanations and their practical use across other prediction domains has also been studied \cite{merrick2020explanation,azeem2024performance}. Our target-specific interpretation therefore identifies the fitted model and its analyzed feature subset explicitly. We also consider prediction-error thresholds and empirical interval behavior as descriptive reliability diagnostics; adaptive uncertainty estimation under distribution shift provides related statistical context \cite{gibbs2021adaptive_conformal}, although the main contribution here is historical-feedback prediction rather than a new uncertainty-quantification algorithm.

\subsection{Research gap and positioning of the present study}
\label{subsec:eaf_related_gap}

A critical challenge in data-driven EAF steelmaking prediction is the development of predictive frameworks that explicitly account for the evolving availability of process measurements and previously observed outcomes. Existing research has primarily examined prediction accuracy, model architectures, and operational feature relationships, while comparatively less attention has been devoted to systematically investigating how historical outcome feedback can be incorporated under prediction-time information constraints. In particular, the predictive contributions of released historical measurements, their temporal availability, and their potential for sequential model adaptation have not been comprehensively characterized within a unified framework addressing both tapping temperature and final steel chemical composition. Addressing this research gap requires moving beyond conventional model-centric comparisons toward an availability-aware learning formulation that integrates historical feedback, chronological evaluation, and adaptive prediction. Accordingly, four interrelated research questions guide the present study:

\begin{enumerate}[label=(\roman*),leftmargin=*,itemsep=2pt]
    \item \textbf{Historical information:} How much predictive value is added by earlier released target measurements beyond current-heat operational and temporal descriptors?
    \item \textbf{Transferability across predictions and algorithms:} Is the effect consistent across tapping temperature, twelve separate chemistry outcomes, and diverse regression families under matched chronological evaluation?
    \item \textbf{Feedback composition and freshness:} Which historical summary groups contribute most, and how does prediction accuracy respond when outcome feedback is assumed to arrive later?
    \item \textbf{Sequential adaptation:} Can released prior prediction errors improve later estimates through a low-overhead residual correction while keeping the fitted regressor fixed?
\end{enumerate}

To investigate these questions, the present study integrates eight industrial source categories into task-specific, cutoff-defined heat records and compares three nested information levels: \textsc{Static}, \textsc{Temporal}, and \textsc{History}. It evaluates twelve regression algorithms under chronological development folds, provides a separate retrospective later-period assessment, and complements the main comparisons with feature-group ablation, delay sensitivity, an exploratory four-architecture DL benchmark, and target-specific attribution diagnostics. The resulting contribution is an \emph{integrated availability-aware prediction formulation and its systematic empirical evaluation} for EAF temperature and elemental composition, rather than the invention of a new regression estimator or feature-importance algorithm.

\section{Industrial EAF Dataset and Prediction Problem Formulation}
\label{sec:eaf_data}

\subsection{Industrial data sources}
\label{subsec:eaf_sources}

This study uses historical operational records from an industrial EAF, described in the source documentation as a DSP-120 installation \cite{iurii_katser_igor_mozolin_maxim_mezhov_vladislav_travnikov_2023}. The heat-indexed archive covers January 2015 through July 2018 and consists of eight heterogeneous data sources: furnace temperature and oxidation measurements, transformer operating records, carbon-injection records, oxygen/gas-lance measurements, basket charging, additional material charging, initial chemical measurements, and final steel chemistry assays. The data combine event-level process histories with intermittently measured quality outcomes. Table~\ref{tab3} summarizes their respective roles in the predictive study.

\begin{table}[!ht]
\centering
\caption{Industrial EAF source records and their roles in the  cutoff-aware prediction protocol. Source-row counts are those documented in the original data inventory; they are not the numbers of eligible modeling heats.}
\label{tab3}
\footnotesize
\begin{tabular}{@{}p{2.95cm}r p{6.70cm}@{}}
\toprule
\textbf{Source file / log} & \textbf{Raw records} & \textbf{Role in analysis} \\
\midrule
EAF temperature and oxidation & 85,104 & Temperature outcome timestamp and value; earlier within-heat temperature and oxidation readings available before the task-specific cutoff. \\
EAF transformer & 271,712 & Completed operating intervals; transformer-stage and duration-based descriptors. \\
Carbon injector usage & 4,011,646 & Pre-cutoff carbon-flow summaries and screened cumulative injection information. \\
Basket charged & 137,685 & Pre-cutoff net basket mass, charging-event times, and material-code diversity. \\
EAF added materials & 79,944 & Pre-cutoff net additions, addition-event times, and material-code diversity. \\
Gas/O$_2$ lance usage & 5,748,194 & Pre-cutoff oxygen and gas-flow summaries, including recent and within-heat temporal descriptors. \\
Initial chemical measurements & 20,827 & Included in the source archive but excluded from the principal operational predictor representations used in the benchmark. \\
EAF final chemical measurements & 3,709 & Recorded outcome measurements for 12 chemical elements; retained by heat and element following target-specific quality screening. \\
\bottomrule
\end{tabular}
\end{table}

\begin{figure}[!ht]
\centering
\includegraphics[width=\linewidth]{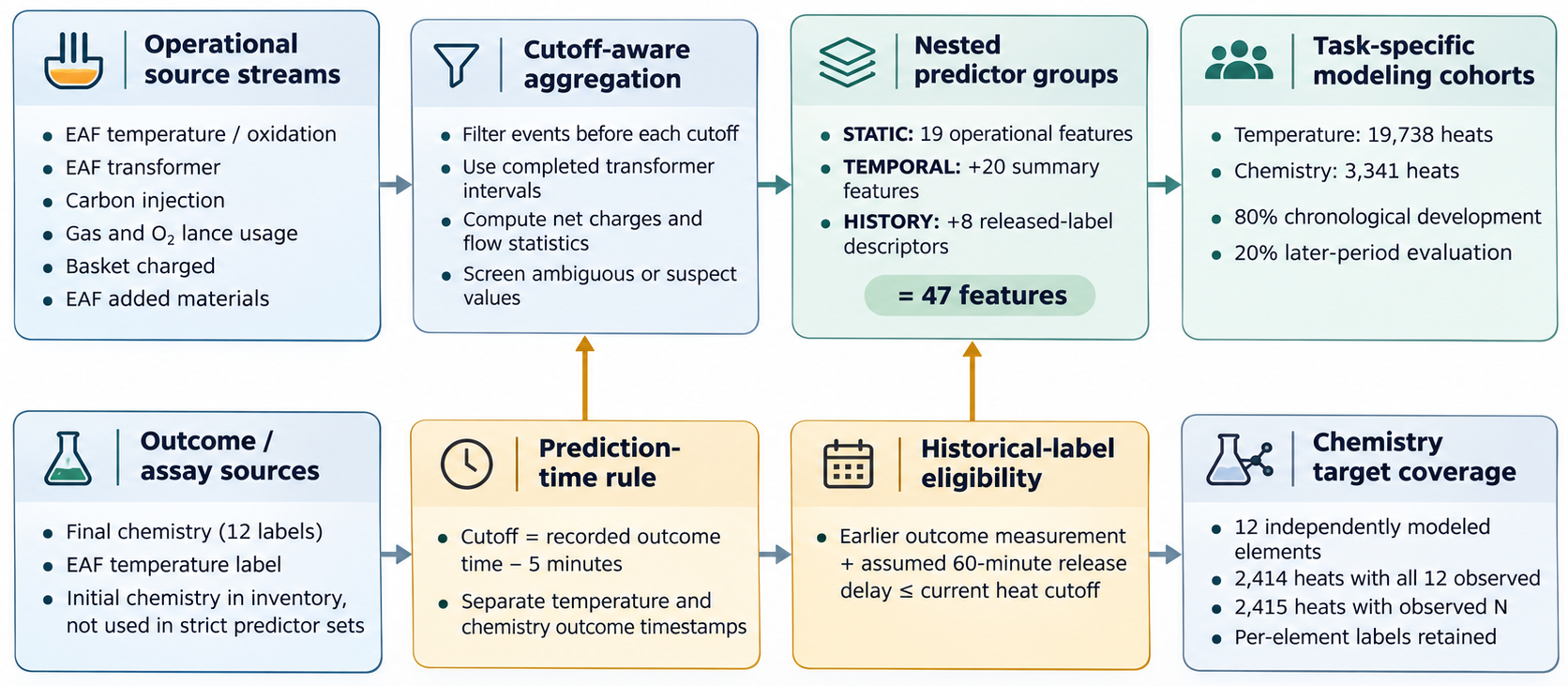}
\caption{Cutoff-aware preprocessing and construction of the two modeling cohorts. Heat-indexed source streams are filtered using task-specific outcome cutoffs, quality-screened, and summarized as operational and within-heat temporal descriptors. Historical outcome features are assembled separately from earlier, assumed-released labels. Final eligible cohorts comprise 19,738 temperature heats and 3,341 chemistry heats; the two cohorts are not required to contain identical heats.}
\label{fig:preprocessing_overview}
\end{figure}

Figure~\ref{fig:preprocessing_overview} outlines the task-specific preprocessing and modeling cohorts. The eligibility and availability rules used to construct this workflow are detailed below.

\subsection{Prediction targets and information-availability cutoffs}
\label{subsec:eaf_cutoff}

Two prediction tasks are considered: (i) the final recorded furnace temperature, expressed in $^\circ$C, and (ii) the recorded final mass fractions of twelve chemical elements, expressed in wt.\%. Let $q\in\mathcal{Q}$ index the thirteen individual prediction targets, where $\mathcal{Q}=\{T,\mathrm{C},\mathrm{Si},\mathrm{Mn},\mathrm{P},\mathrm{S},\mathrm{Cu},\mathrm{Cr},\mathrm{Mo},\mathrm{Ni},\mathrm{As},\mathrm{Sn},\mathrm{N}\}$. Here, $T$ denotes tapping temperature, and the remaining twelve indices denote individual chemical elements. The observed value of target $q$ for heat $i$ is denoted by $y_i^{(q)}$. A heat is the unit of analysis, identified by a unique \texttt{HEATID}. For each target, the endpoint value is taken from the last unambiguous valid reading at its recorded measurement timestamp; conflicting measurements at the same last timestamp are treated as unresolved rather than arbitrarily selected.

To prevent the use of contemporaneous endpoint measurements or process records observed after the prediction point, the preprocessing defines a task-specific historical prediction cutoff,
\begin{equation}
 t_{i,\mathrm{cut}}^{(q)} = t_{i,\mathrm{obs}}^{(q)} - \Delta,
\qquad q\in\mathcal{Q},\quad \Delta=5\ \mathrm{min},
 \label{eq:eaf_cutoff}
\end{equation}
where $t_{i,\mathrm{obs}}^{(q)}$ denotes the recorded measurement timestamp for target $q\in\mathcal{Q}$. Event measurements are used only if their timestamps precede the corresponding cutoff; transformer intervals must be fully completed before that cutoff. The two task-specific cutoffs need not coincide for a given heat. The five-minute lead is a retrospective prediction-time convention based on the available record timestamps, rather than an independently verified interval before physical tapping.

\begin{figure}[!ht]
\centering
\includegraphics[width=\linewidth]{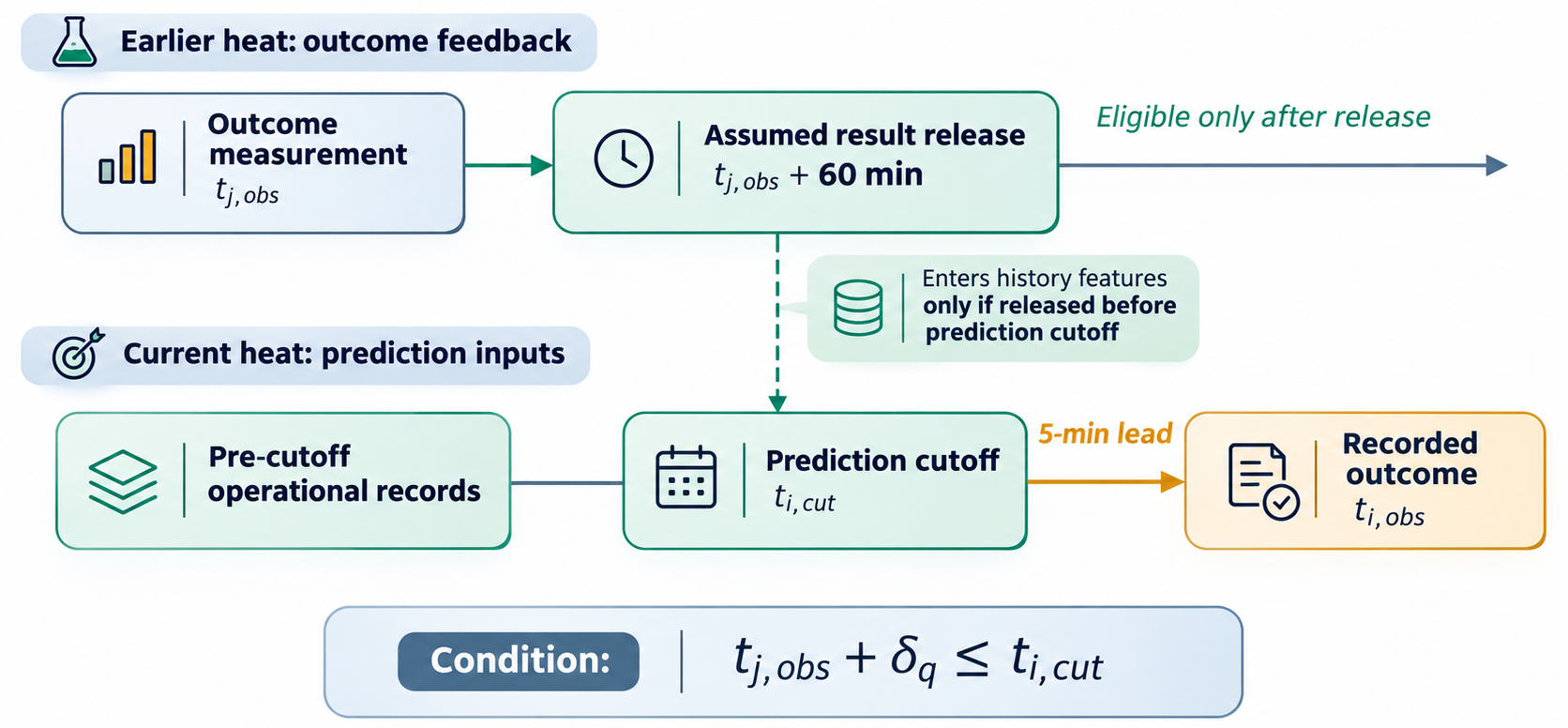}
\caption{Prediction-time and historical-outcome availability assumptions. The task-specific prediction cutoff occurs five minutes before the corresponding recorded outcome measurement. Earlier temperature or composition outcomes become eligible as historical inputs only after the assumed release delay; physical tapping and laboratory reporting times were not independently verified.}
\label{fig:eaf_availability}
\end{figure}

For historical-outcome feedback, an earlier outcome $j$ is considered eligible for prediction $i$ only if
\begin{equation}
 t_{j,\mathrm{obs}}^{(q)} + \delta_q \le t_{i,\mathrm{cut}}^{(q)},
 \label{eq:eaf_label_release}
\end{equation}
where $\delta_q=60\,\mathrm{min}$ denotes the prespecified outcome-release delay adopted in the principal experiments. Under this scenario, an earlier outcome becomes eligible for historical feedback 60 minutes after its recorded measurement timestamp, ensuring that only previously released measurements contribute to the current prediction. Since actual laboratory-reporting and label-release timestamps were unavailable, alternative release-delay scenarios are also investigated through sensitivity analysis. Figure~\ref{fig:eaf_availability} illustrates the distinction between current-heat operational information and previously released outcomes.

\subsection{Cutoff-aware preprocessing and heat-level assembly}
\label{subsec:eaf_preprocessing}

A comprehensive, cutoff-aware data-engineering procedure was developed to transform the heterogeneous industrial records into consistent heat-level predictors. The preprocessing involved timestamp and heat-identifier harmonization, numerical validation, prediction-time filtering, and source-specific feature construction. Temperature and oxidation measurements were restricted to valid pre-cutoff observations, while transformer descriptors were calculated from completed operating intervals, including stage information and the union of operating durations to prevent double-counting overlapping records. The recorded transformer power/energy field was excluded from the principal feature set because its physical interpretation could not be reliably established from the available metadata.

Charging and material-addition records were processed using signed masses to determine net material inputs, together with distinct event and material-code counts. Carbon, oxygen, and gas-injection records were transformed into time-weighted flow statistics, peak values, and within-heat temporal descriptors. Quality screening addressed inconsistent cumulative counters, suspicious oxidation readings of 9,999 ppm, and ambiguous target measurements. Missing predictors were handled through training-partition-specific imputation, preventing information from subsequent validation periods from entering model preparation. These procedures establish reliable, physically interpretable input representations while preserving the information available at each prediction cutoff.

Figure~\ref{fig:preprocessing_overview} summarizes the cutoff-aware processing of heterogeneous industrial records into task-specific modeling cohorts. Operational measurements are filtered according to each task's prediction cutoff before aggregation and heat-level integration, while subsequent outcome measurements are retained as prediction targets. This procedure establishes a consistent separation between information available at prediction time and the outcomes being estimated.

\subsection{Static, temporal, and historical-feedback representations}
\label{subsec:eaf_features}

To distinguish the information gained from within-heat temporal behavior and from previously released outcome feedback, three nested predictor configurations were constructed for both tasks: \emph{STATIC} (19 predictors), \emph{TEMPORAL} (the 19 static predictors plus 20 time-resolved aggregate descriptors), and \emph{HISTORY} (the 39 temporal predictors plus eight availability-aware historical-outcome descriptors). Table~\ref{tab:final_features} specifies the component groups and feature counts used in the prediction experiments.

\begin{table}[!ht]
\centering
\caption{Predictor groups used in the EAF prediction benchmark. Historical descriptors are computed separately for the relevant temperature or elemental-composition outcome; \texttt{HEATID}, measurement times, cutoff timestamps, and outcome labels are not model predictors.}
\label{tab:final_features}
\footnotesize
\begin{tabular}{@{}p{3.25cm}c p{7.95cm}@{}}
\toprule
\textbf{Group} & \textbf{No.} & \textbf{Feature definitions} \\
\midrule
Pre-cutoff temperature / oxidation (STATIC) & 2 & Latest earlier temperature; latest earlier oxidation measurement. \\
Transformer (STATIC) & 4 & Completed interval count; maximum recorded stage; number of distinct stages; union duration of completed intervals. \\
Basket charging (STATIC) & 3 & Net charged mass; number of distinct charging timestamps; number of distinct material codes. \\
Additional materials (STATIC) & 3 & Net added mass; number of distinct addition timestamps; number of distinct material codes. \\
O$_2$, gas, and carbon usage (STATIC) & 7 & Time-weighted means and peaks of three flow streams (six); screened carbon counter total (one). \\
\midrule
\textbf{STATIC total} & \textbf{19} & \textbf{Operational information available before the task-specific cutoff.} \\
\midrule
Recent-window behavior (TEMPORAL addition) & 8 & Completed transformer union duration and record count in the last 10 minutes; last-10-minute mean and peak for O$_2$, gas, and carbon flows. \\
Within-heat flow stages (TEMPORAL addition) & 12 & First-, middle-, and last-third flow medians and last-minus-first contrast, for each of O$_2$, gas, and carbon. \\
\midrule
\textbf{TEMPORAL total} & \textbf{39} & \textbf{STATIC plus 20 temporal summaries.} \\
\midrule
Released-outcome history (HISTORY addition) & 8 & Last released value; second-last released value; averages over 5 and 20 released values; standard deviation over 20 released values; minutes since last released measurement; count released in 24 hours; short-minus-long moving-average contrast. \\
\midrule
\textbf{HISTORY total} & \textbf{47} & \textbf{TEMPORAL plus eight feedback descriptors.} \\
\bottomrule
\end{tabular}
\end{table}

The model input for heat $i$ is therefore $\mathbf{x}_i^{(S)}\in\mathbb{R}^{19}$, $\mathbf{x}_i^{(T)}\in\mathbb{R}^{39}$, or $\mathbf{x}_i^{(H)}\in\mathbb{R}^{47}$, subject to missing values handled within each training fold. Each chemistry element is predicted with its own fitted regressor; sharing feature definitions does not imply that dependencies among the twelve response variables are explicitly modeled. The construction and sequential updating of the historical-feedback descriptors are further described in the proposed-framework section.

\subsection{Final modeling cohorts and chemical-assay coverage}
\label{subsec:eaf_cohorts}

Following target-specific quality screening and prediction-time alignment, the temperature and chemical-composition datasets contain \textbf{19,738} and \textbf{3,341} eligible heats, respectively. Each prediction task uses its corresponding eligible observations and chronological ordering. The earliest 80\% comprise the development datasets, containing 15,790 temperature heats and 2,672 chemistry heats, while the remaining 3,948 and 669 heats form the retrospective later-period evaluation datasets. Chronological development validation and later-period evaluation are described separately in Section~\ref{sec:eaf_experiments}.

Chemical-composition labels are retained \emph{per element}, rather than discarding a heat whenever any of the twelve assays is missing. Table~\ref{tab:eaf_target_coverage} reports the resulting target-specific availability. Nitrogen has substantially lower coverage than the other elements (2,415 of 3,341 heats, or 72.3\%). Only 2,414 heats have all twelve quality-screened targets simultaneously observed (72.3\% of the chemistry cohort), so restricting all model fits to complete cases would unnecessarily discard useful observations for the well-measured elements.

\begin{table}[!ht]
\centering
\caption{Availability of quality-screened endpoint chemistry targets within the 3,341-heat composition cohort, including the annual distribution of nitrogen assay availability. Counts are reported after element-specific target screening and before chronological partitioning.}
\label{tab:eaf_target_coverage}
\footnotesize
\begin{tabular}{@{}lrrr@{}}
\toprule
\textbf{Element} & \textbf{Available heats} & \textbf{Missing heats} & \textbf{Available (\%)} \\
\midrule
C  & 3,340 & 1 & 99.97 \\
Si & 3,341 & 0 & 100.00 \\
Mn & 3,341 & 0 & 100.00 \\
P  & 3,340 & 1 & 99.97 \\
S  & 3,341 & 0 & 100.00 \\
Cu & 3,341 & 0 & 100.00 \\
Cr & 3,341 & 0 & 100.00 \\
Mo & 3,341 & 0 & 100.00 \\
Ni & 3,341 & 0 & 100.00 \\
As & 3,341 & 0 & 100.00 \\
Sn & 3,340 & 1 & 99.97 \\

N & 2,415 & 926 & 72.28 \\
\midrule
\multicolumn{4}{@{}l}{\emph{Nitrogen availability by production year}} \\
N (2015) & 518 & 20 & 96.28 \\
N (2016) & 672 & 96 & 87.50 \\
N (2017) & 290 & 101 & 74.17 \\
N (2018) & 935 & 709 & 56.87 \\
\bottomrule
\end{tabular}
\end{table}

Nitrogen assay coverage varied across production years, decreasing from 96.28\% in 2015 to 87.50\% in 2016, 74.17\% in 2017, and 56.87\% in 2018, as summarized in Table~\ref{tab:eaf_target_coverage}. Similarly, coverage was 96.86\% in the earliest development block, 69.16\% across development-validation folds, and 57.10\% in the later-period evaluation. The element-wise modeling approach preserves available observations for other chemical targets when nitrogen measurements are missing. Accordingly, nitrogen prediction performance is reported using observed-label subsets, with temporal differences in assay availability considered when interpreting the results.

\section{Proposed Availability-Aware Historical-Feedback Framework}
\label{sec:eaf_framework}

\subsection{Framework overview}
\label{subsec:eaf_framework_overview}

The proposed framework treats EAF endpoint prediction as a sequence of heat-indexed regression problems in which the information available at the prediction instant differs from the information available after the heat has been completed. For each eligible heat $i$ and prediction target $q$, a model uses (i) operational measurements recorded before the task-specific prediction cutoff and (ii) outcomes of earlier heats that are considered available by that cutoff. The target index $q\in\mathcal{Q}$ follows the definition in Section~\ref{subsec:eaf_cutoff} and represents either tapping temperature or one of the twelve individually modeled final chemical-composition targets. Separate models are fitted for each target; no explicit dependency among different chemical responses is imposed.

\begin{figure}[!ht]
    \centering
    \includegraphics[width=\linewidth]{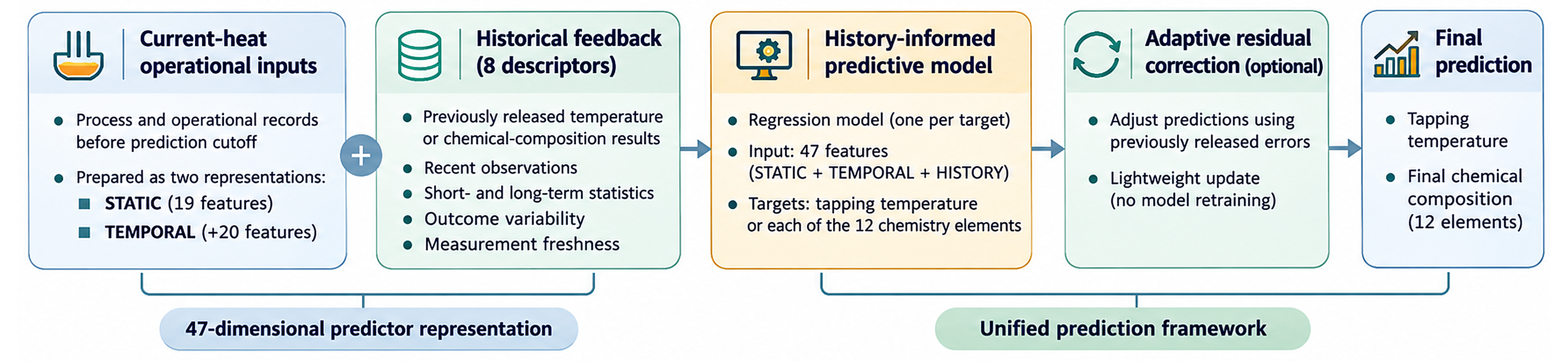}
    \caption{Architecture of the history-informed EAF prediction framework. Pre-cutoff operational data are represented at three information levels, while only eligible earlier outcomes contribute to the history descriptors. Each target is predicted by an independently fitted regressor. A separately defined adaptive correction uses released errors of earlier \emph{base-model predictions}; those errors are not members of the eight historical-outcome descriptors.}
    \label{fig:eaf_s4_framework}
\end{figure}

Figure~\ref{fig:eaf_s4_framework} summarizes the distinction between the main history-informed predictor and the adaptive online residual-correction layer. The core representation is independent of the regression algorithm: the same cutoff-defined input vectors can be used with linear, instance-based, kernel, or ensemble regressors. Consequently, the experimental analysis can distinguish improvements attributable to historical information from differences attributable to model choice. Data sources, endpoint timestamps, and the task-specific cutoff construction are defined in Section~\ref{sec:eaf_data}; the chronological model-training and evaluation protocol is specified in the following experimental-design section.

\subsection{Nested heat-level predictor representations}
\label{subsec:eaf_framework_representations}

For heat $i$, let $\mathbf{s}_i\in\mathbb{R}^{19}$ contain the static operational descriptors and let $\mathbf{a}_i\in\mathbb{R}^{20}$ contain the additional within-heat temporal summaries. Let $\mathbf{h}_i^{(q)}\in\mathbb{R}^{8}$ denote the historical-outcome descriptors computed for target $q$. The three nested input configurations are
\begin{equation}
\begin{aligned}
 \mathbf{x}_i^{(S)} &= \mathbf{s}_i,\\
 \mathbf{x}_i^{(T)} &= [\mathbf{s}_i^{\mathsf T},\,\mathbf{a}_i^{\mathsf T}]^{\mathsf T},\\
 \mathbf{x}_i^{(H,q)} &= [\mathbf{s}_i^{\mathsf T},\,\mathbf{a}_i^{\mathsf T},\,(\mathbf{h}_i^{(q)})^{\mathsf T}]^{\mathsf T},
\end{aligned}
\label{eq:eaf_s4_nested_inputs}
\end{equation}
with respective dimensions of 19, 39, and 47. Here, \emph{temporal} refers to summaries of within-heat measurements recorded prior to the cutoff, whereas \emph{historical} refers to outcomes from earlier heats. In particular, the history features for a composition target use previously released measurements of that \emph{same} element; they do not introduce the current heat's final assay or other current-heat endpoint measurements as predictors.

For a given predictor configuration $c\in\{S,T,H\}$, the base regression estimate is
\begin{equation}
 \widehat y_{i,\mathrm{base}}^{(q,c)}
   = f_{q,c}\!\left(\mathbf{x}_i^{(c,q)};\,\widehat{\boldsymbol\theta}_{q,c}\right),
\label{eq:eaf_s4_base_model}
\end{equation}
where the fitted model parameters and the selected hyperparameters $\widehat{\boldsymbol\theta}_{q,c}$ are obtained from the appropriate chronological development data. For $c=S$ or $T$, the feature vector is common across targets; for $c=H$, it is target-specific. Feature values that cannot be computed because insufficient history is available are left missing and handled by preprocessing fitted on the corresponding training partition, rather than filled using later observations.

\subsection{Release-aware historical-outcome descriptors}
\label{subsec:eaf_framework_history}

The primary prediction experiments adopt the task-specific cutoff $t_{i,\mathrm{cut}}^{(q)}$ and the assumed release-delay rule defined in Eqs.~\eqref{eq:eaf_cutoff} and \eqref{eq:eaf_label_release}. For target $q$, define the eligible historical set at heat $i$ as
\begin{equation}
 \mathcal{H}_i^{(q)} =
 \left\{(y_j^{(q)},r_j^{(q)}):
   r_j^{(q)}=t_{j,\mathrm{obs}}^{(q)}+\delta_q
   \leq t_{i,\mathrm{cut}}^{(q)},\;
   y_j^{(q)}\text{ is observed}\right\},
\label{eq:eaf_s4_available_set}
\end{equation}
where $\delta_q=60\,$min in the primary availability scenario. Entries are ordered by their assumed release timestamps $r_j^{(q)}$; accordingly, a historical summary cannot incorporate a measurement whose assumed release is later than the current prediction cutoff. The 60-minute interval is an explicit retrospective availability assumption rather than a recorded plant reporting time. The separate delay-sensitivity experiments assess alternative assumptions.

Write the eligible outcomes, in release order, as $z_{i,1}^{(q)},\ldots,z_{i,n_i}^{(q)}$, where $n_i=|\mathcal H_i^{(q)}|$, and define $n_{i,m}=\min(m,n_i)$. When $n_i>0$, the rolling mean of the last $m$ released values is
\begin{equation}
 \mu_{i,m}^{(q)}=\frac{1}{n_{i,m}}
 \sum_{\ell=n_i-n_{i,m}+1}^{n_i}z_{i,\ell}^{(q)},
 \qquad m\in\{5,20\}.
\label{eq:eaf_s4_history_mean}
\end{equation}
The eight descriptors are collected in
\begin{equation}
 \mathbf{h}_i^{(q)}=
 \bigl[\;z_{i,n_i}^{(q)},\ z_{i,n_i-1}^{(q)},\
 \mu_{i,5}^{(q)},\ \mu_{i,20}^{(q)},\
 \sigma_{i,20}^{(q)},\ \tau_i^{(q)},\
 N_{i,24\mathrm{h}}^{(q)},\ \mu_{i,5}^{(q)}-\mu_{i,20}^{(q)}\;\bigr]^{\mathsf T},
\label{eq:eaf_s4_history_vector}
\end{equation}
where $\sigma_{i,20}^{(q)}$ is the population-standard-deviation estimate of the last $n_{i,20}$ released values (defined when at least two values exist), $\tau_i^{(q)}=t_{i,\mathrm{cut}}^{(q)}-\max_{j\in\mathcal H_i^{(q)}}r_j^{(q)}$ is the elapsed time since the most recent \emph{release}, expressed in minutes, and $N_{i,24\mathrm{h}}^{(q)}$ counts eligible releases during the preceding 24 hours. The first two entries are the most recent and second-most-recent released outcomes, respectively. The difference between the short- and long-window averages represents the local direction of the outcome history. When there is no eligible outcome, the count is zero and the remaining unavailable summaries are marked missing; the second-most-recent value and standard deviation are also missing until enough observations exist.

This construction explicitly separates information from within the current heat from outcomes of preceding heats. Importantly, it uses observed historical \emph{target values} rather than residuals of previous predictions. Errors of previous predictions enter only the adaptive residual-correction mechanism described next.

\subsection{Lightweight online residual correction}
\label{subsec:eaf_framework_adaptive}

To test whether feedback can additionally compensate for locally persistent prediction bias, a lightweight additive adjustment is applied to the history-informed base model without modifying its fitted parameters. Let $\widehat y_{j,\mathrm{base}}^{(q)}$ denote the base prediction saved for an earlier heat $j$. Once its true outcome has become eligible, the corresponding \emph{base-model} residual is
\begin{equation}
 e_j^{(q)}=y_j^{(q)}-\widehat y_{j,\mathrm{base}}^{(q)}.
\label{eq:eaf_s4_residual}
\end{equation}
For a prediction at cutoff $t_{i,\mathrm{cut}}^{(q)}$, define $\mathcal E_i^{(q)}$ as the most recent $L_q$ previously predicted cases whose residuals are available because their target outcomes have been released by that cutoff. If at least three such cases exist, their weighted mean error is
\begin{align}
 \overline e_i^{(q)} & =
 \frac{\displaystyle\sum_{j\in\mathcal E_i^{(q)}}w_{ij}^{(q)}e_j^{(q)}}
      {\displaystyle\sum_{j\in\mathcal E_i^{(q)}}w_{ij}^{(q)}},
 \label{eq:eaf_s4_weighted_residual}\\
 w_{ij}^{(q)} & =\begin{cases}
 1, & \text{uniform weighting},\\[2pt]
 \exp\!\left[-\dfrac{\ln 2}{h_q}\bigl(t_{i,\mathrm{cut}}^{(q)}-r_j^{(q)}\bigr)\right],
 & \text{exponential weighting}.
 \end{cases}\notag
\end{align}
where $h_q$ is a time-based half-life when exponential weighting is used. The corrected prediction is
\begin{equation}
 \widehat y_{i,\mathrm{adapt}}^{(q)}
 =\widehat y_{i,\mathrm{base}}^{(q,H)}+\Delta_i^{(q)},
 \quad
 \Delta_i^{(q)}=
 \begin{cases}
 \lambda_q\overline e_i^{(q)}, & |\mathcal E_i^{(q)}|\geq3,\\
 0, & |\mathcal E_i^{(q)}|<3.
 \end{cases}
\label{eq:eaf_s4_adaptive_prediction}
\end{equation}
The nonnegative correction strength $\lambda_q$, memory length $L_q$, and the selected recency-weighting scheme are chosen using chronological out-of-fold predictions from the development period. In the implementation, the candidate strengths are $\{0,0.25,0.5,1\}$, memory lengths are $\{5,20,50\}$, and the weighting alternatives are uniform or exponential with a 48-hour half-life. Correction parameters are selected using the first half of the final development-validation fold, without using later-period outcomes for parameter selection.

\begin{figure}[!ht]
    \centering
    \includegraphics[width=\linewidth]{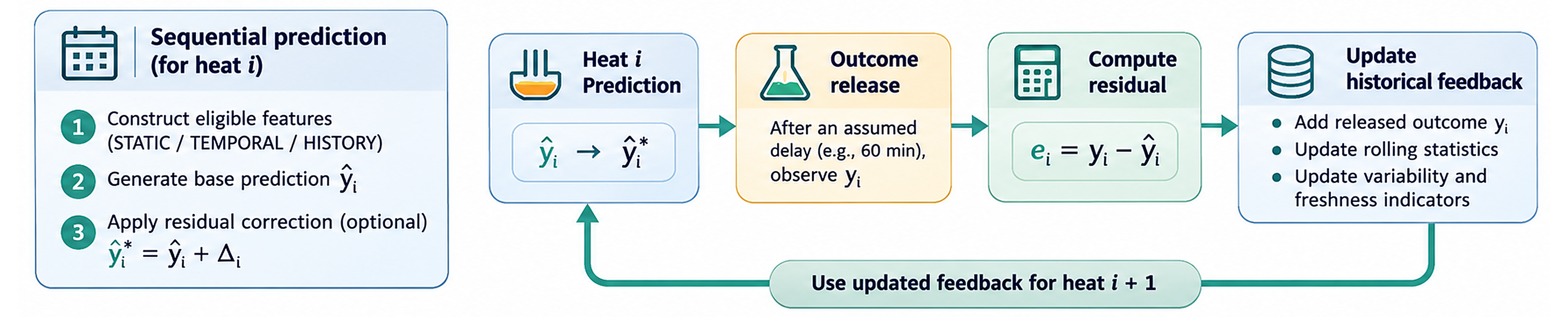}
    \caption{Sequential residual-feedback adjustment for a target-specific frozen model. Earlier observed outcomes become eligible after their assumed release, allowing stored \emph{base-model} prediction errors to be summarized and used for an adaptive additive correction. After the current outcome is subsequently released, its base-model residual can enter memory for later heats; current and unreleased outcomes are never available to the current prediction.}
    \label{fig:eaf_s4_adaptive}
\end{figure}

The adjustment is deliberately distinguished from model retraining: the regression function $f_{q,H}$ and its trained parameters remain frozen, while its input history and the residual-feedback memory evolve as additional labels are assumed to be released. The implemented later-period evaluation initializes residual memory within that evaluation sequence, so initial cases without three previously released residuals receive the unadjusted base prediction. Figure~\ref{fig:eaf_s4_adaptive} illustrates the two feedback paths without implying that residuals are members of the eight history descriptors.

\subsection{Sequential inference and updating procedure}
\label{subsec:eaf_framework_sequential}

For each target, the prediction sequence is evaluated in chronological cutoff order. At each heat $i$, the inference procedure: (i) obtains current-heat static and temporal measurements restricted to the task-specific cutoff; (ii) determines which earlier observed target values have become eligible under Eq.~\eqref{eq:eaf_s4_available_set}; (iii) computes the eight historical descriptors and evaluates the frozen history-informed regression model; and, if adaptive correction is enabled, (iv) adjusts the base prediction using only previously released base-model residuals. The current heat's prediction and cutoff timestamp are saved. Its outcome and associated residual are added to the respective memories only after the assumed release time, so they may contribute to \emph{subsequent} predictions but not to their own prediction.

This prequential procedure allows later predictions within a chronological validation or evaluation segment to use outcomes from earlier cases \emph{once those outcomes are eligible}, while keeping the regression parameters fixed within each fitted evaluation model. The resulting comparison therefore tests the contribution of available feedback under a retrospective timing scenario, rather than assuming access to all endpoint labels at the start of a validation block. The forward-validation design, preprocessing fitted within training partitions, expanded hyperparameter searches, baseline comparisons, and evaluation metrics are described in the experimental-design section.

\section{Experimental Design and Evaluation Protocol}
\label{sec:eaf_experiments}

\subsection{Experimental objectives and prediction configurations}
\label{subsec:eaf_exp_overview}

The experiments evaluate three related questions: whether information from earlier, available heat outcomes improves predictions beyond the operational measurements of the current heat; whether the improvement persists across regression algorithms, target elements and feedback-release assumptions; and whether a frozen history-informed model can be adjusted using subsequently released prediction errors. Temperature and final chemical composition are evaluated as separate prediction tasks. The composition task consists of twelve independent, element-specific regressions with separately observed labels. For both tasks, the \textsc{Static}, \textsc{Temporal}, and \textsc{History} predictor configurations contain 19, 39, and 47 predictors, respectively, as defined in Sections~\ref{subsec:eaf_cutoff}--\ref{subsec:eaf_framework_history}. All primary model comparisons use identical task-specific heat partitions and feature-availability rules within each target, permitting paired evaluation of predictor configurations.

\begin{figure}[!ht]
    \centering
    \includegraphics[width=\linewidth]{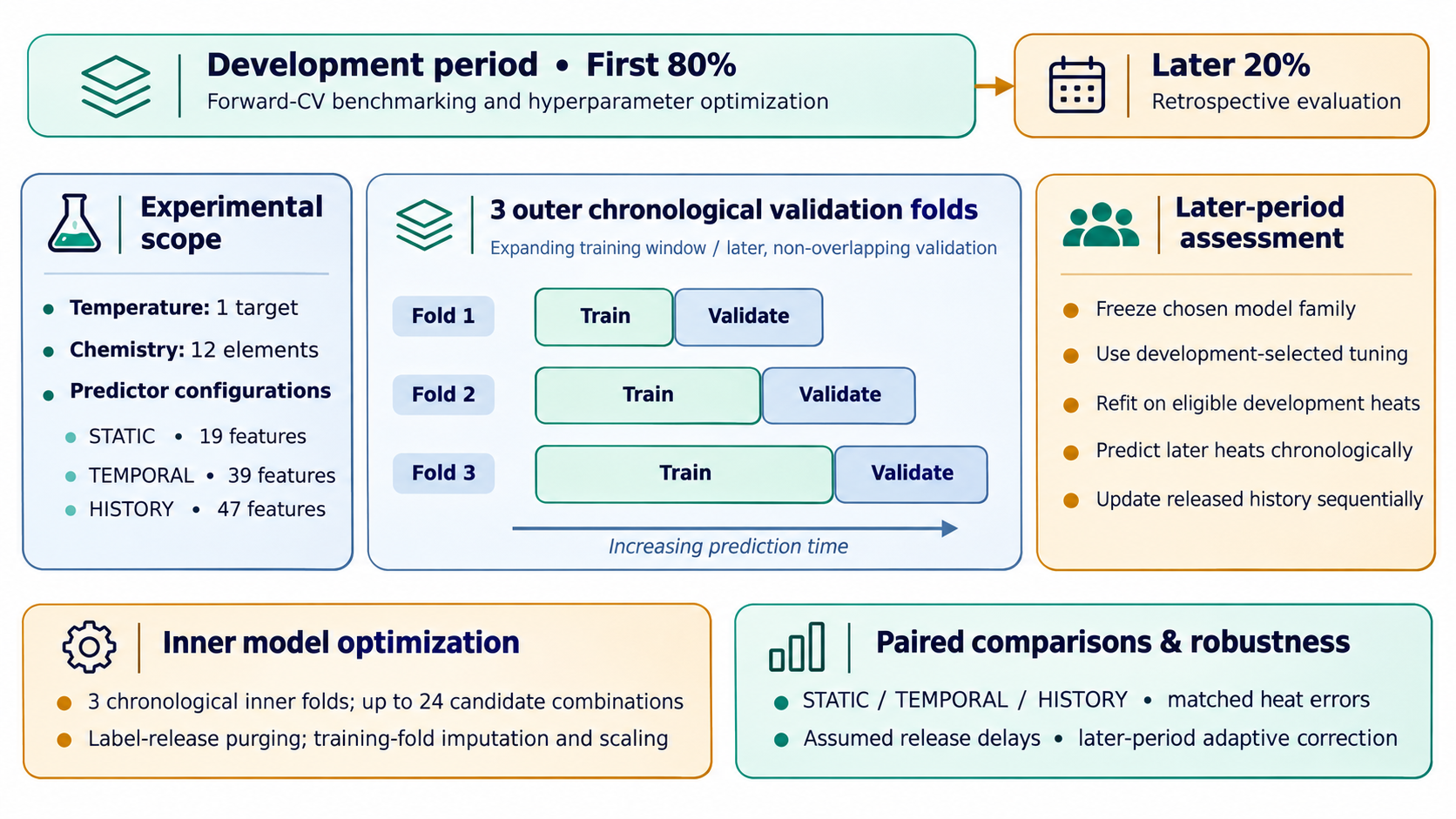}
    \caption{Chronological model-development and retrospective evaluation protocol. The first 80\% of each task-specific, cutoff-sorted heat sequence is used for three expanding-window outer validation folds. Each outer training partition contains a three-fold chronological hyperparameter search; the outer validation outcomes do not determine the fitted parameters for that fold. A selected model is subsequently refitted on eligible development heats for evaluation on the later 20\% of the sequence. The later-period analysis is retrospective and does not constitute a previously unseen external dataset.}
    \label{fig:eaf_s5_protocol}
\end{figure}

The evaluation comprises (i) expanded model benchmarking within the chronological development period, (ii) availability-aware baselines and historical-feedback comparisons, (iii) focused historical-feature ablation and outcome-release delay sensitivity, (iv) retrospective later-period evaluation with adaptive residual correction, (v) an exploratory comparison with four DL architectures across all thirteen prediction targets, and (vi) target-specific model interpretation using SHAP. The overall protocol is illustrated in Figure~\ref{fig:eaf_s5_protocol} and summarized in Table~\ref{tab:eaf_exp_settings}. Development-period forward-validation results and later-period retrospective results are reported separately. The latter-period outcomes were not used in the hyperparameter searches described below.

\subsection{Chronological partitioning and nested model evaluation}
\label{subsec:eaf_exp_cv}

For each task, heats were ordered by their task-specific prediction-cutoff timestamp, using the heat identifier to resolve ties. The earliest 80\% of the resulting heat sequence formed the \emph{development period}; the remaining 20\% formed the later-period retrospective evaluation interval. The two tasks were partitioned independently because their eligible heat cohorts and endpoint timestamps differ. The temperature cohort contains 19,738 heats and the chemistry cohort contains 3,341 heats before the temporal split. In the development period, three expanding-window outer splits were generated using \texttt{TimeSeriesSplit(n\_splits=3)}. Each subsequent fold enlarged the available training history while testing on later, non-overlapping consecutive heats. The fold boundaries were defined from \emph{all} eligible chemistry heats, before target-specific missing-label filtering, so that the same calendar ordering was used for all twelve composition targets.

To avoid training on target values that would not yet have become available at the beginning of an evaluation segment, each outer training set was further purged of observations with assumed target-release times later than the first validation cutoff. Analogous purging was performed in the inner tuning folds. The validation sets contained only heats with observed quality-screened labels for the relevant target, while predictor construction used the complete eligible sequence. Within validation, historical descriptors were computed prequentially: an earlier validation heat's outcome could influence \emph{later} predictions only after its assumed release time. This simulates sequential outcome availability while leaving each fold's fitted regression parameters unchanged. The primary release delay was 60\,min, as specified in Section~\ref{subsec:eaf_cutoff}; it is a declared retrospective assumption rather than a recorded laboratory turnaround time.

For each outer training period, hyperparameters were optimized using three inner chronological validation folds, with mean inner-fold RMSE serving as the selection criterion. The selected estimator was then refitted on the eligible outer-training observations and evaluated on the subsequent validation fold. This nested procedure separates hyperparameter optimization from the chronological evaluation of each regression algorithm. Overall algorithm rankings were subsequently determined from pooled development-validation predictions, while the contribution of historical feedback was assessed through matched comparisons within the same algorithm and prediction target.

\subsection{Candidate algorithms, search spaces, and preprocessing}
\label{subsec:eaf_exp_tuning}

Twelve established regression algorithms were compared: Ridge, Lasso, ElasticNet, Bayesian Ridge, Decision Tree, $k$-Nearest Neighbors (KNN), support vector regression (SVR), AdaBoost, Gradient Boosting, Random Forest, Extremely Randomized Trees (ExtraTrees), and Extreme Gradient Boosting (XGBoost). The algorithms include established tree-ensemble formulations for Random Forest and ExtraTrees \cite{cutler2012random,geurts2006extremely}, gradient-based boosting methods \cite{natekin2013gradient,chen2016xgboost}, and instance-based or kernel regression families \cite{steinbach2009knn,awad2015support}. The parameter grids are listed in Table~\ref{tab:hyperparams}. A reproducible set of at most 24 distinct parameter configurations was evaluated for each algorithm, target, feature configuration, and outer fold. If a candidate grid contained at most 24 combinations, all were tested; otherwise, 24 combinations were sampled without replacement using the prescribed random seed. This protocol is an expanded, budgeted hyperparameter search rather than an exhaustive search over the full grids. The main experiment used random seed 2026, three outer folds and three inner folds.

\begin{table}[!ht]
\caption{Hyperparameter search spaces used in the expanded twelve-algorithm benchmark. The final column reports the number of configurations evaluated in each outer training fold; ranges are discrete candidate values, not continuous intervals.}
\label{tab:hyperparams}
\centering\footnotesize
\setlength{\tabcolsep}{4pt}
\renewcommand{\arraystretch}{1.10}
\begin{tabular}{@{}p{2.55cm}p{9.0cm}r@{}}
\toprule
\textbf{Algorithm} & \textbf{Tuned hyperparameters and candidate values} & \textbf{Tests} \\
\midrule
Ridge & $\alpha\in\{0.01,0.1,1,10,50,100\}$ & 6 \\
Lasso & $\alpha\in\{0.0001,0.0003,0.001,0.003,0.01,0.03,0.08\}$ & 7 \\
Elastic Net & $\alpha\in\{0.0001,0.001,0.01,0.03,0.1\}$; $\ell_1$ ratio $\in\{0.2,0.5,0.8,1\}$ & 20 \\
Bayesian Ridge & $\alpha_1,\alpha_2\in\{10^{-6},10^{-5},10^{-4}\}$ & 9 \\
Decision Tree & Maximum depth $\in\{\mathrm{None},5,10,20\}$; minimum split $\in\{2,5,10\}$; minimum leaf $\in\{1,3\}$ & 24 \\
KNN & Neighbors $\in\{3,5,7,9,15,21\}$; weights $\in\{\mathrm{uniform},\mathrm{distance}\}$ & 12 \\
SVR & $C\in\{0.1,1,10\}$; $\varepsilon\in\{0.01,0.1,0.5\}$; kernel $\in\{\mathrm{RBF},\mathrm{polynomial}\}$ & 18 \\
AdaBoost & Estimators $\in\{50,100,200\}$; learning rate $\in\{0.01,0.1,1\}$ & 9 \\
Gradient Boosting & Estimators $\in\{100,200,500\}$; learning rate $\in\{0.01,0.05,0.1,0.2\}$; maximum depth $\in\{3,5,7\}$ & 24 \\
Random Forest & Estimators $\in\{100,200,500\}$; maximum depth $\in\{\mathrm{None},10,20\}$; minimum split $\in\{2,5,10\}$; maximum features $\in\{0.5,0.75,1\}$ & 24 \\
ExtraTrees & Estimators $\in\{100,200,500\}$; maximum depth $\in\{\mathrm{None},10,20\}$; minimum split $\in\{2,5,10\}$; maximum features $\in\{0.5,0.75,1\}$ & 24 \\
XGBoost & Estimators $\in\{100,200,500\}$; learning rate $\in\{0.01,0.05,0.1\}$; maximum depth $\in\{3,5,7\}$; row/column sampling $\in\{0.7,0.85,1\}$ & 24 \\
\bottomrule
\end{tabular}
\end{table}

All models used a pipeline in which missing input values were imputed using medians estimated from the fitting data alone. Predictors were standardized only for Ridge, Lasso, ElasticNet, Bayesian Ridge, KNN, and SVR; transformations were fitted within the corresponding chronological training partitions. Because the twelve composition targets have markedly different numerical scales, the target variable was additionally standardized \emph{within the fitted regression pipeline} for Lasso, ElasticNet, and SVR and inversely transformed for prediction. These preprocessing choices were part of inner-model fitting rather than operations performed on the full dataset before splitting. The candidate search spaces and evaluation settings are summarized in Table~\ref{tab:hyperparams}, while hyperparameter selection is performed independently within the corresponding chronological training folds.

\subsection{Benchmark baselines and controlled feature comparisons}
\label{subsec:eaf_exp_baselines}

The three feature configurations provide a controlled within-algorithm comparison: \textsc{Static} versus \textsc{Temporal} isolates the effect of adding within-heat temporal summaries, while \textsc{Temporal} versus \textsc{History} isolates the additional information from eligible earlier target outcomes. This is preferable to inferring the value of historical features from a comparison between \emph{different} model families. Two no-learning reference predictors were also computed on matching chronological validation heats. The \emph{outer-training median} uses the target median from the eligible outer training records. The \emph{last-20-released median} uses the median of the 20 most recently released values of the same target at each prediction cutoff, reverting to the training median if no earlier target outcome is eligible. The latter baseline may incorporate earlier validation outcomes once they have been released, consistently with the sequential availability rule.

A focused history-group ablation was conducted on the third development validation fold, with the regression hyperparameters carried over from that fold's tuned \textsc{History} model. The configurations were: A0, temporal predictors alone; A1, A0 plus the last released outcome; A2, A1 plus the mean of the last five released outcomes; A3, A2 plus the mean and standard deviation of the last twenty released outcomes; and A4, the complete eight-descriptor historical representation. The last configuration reused the benchmark predictions, and the other configurations required only a small number of refits. This ablation was conducted for the selected temperature model and representative chemistry targets P, Mo, Si, and N, and was interpreted as a focused development-fold mechanism study rather than a comprehensive independent validation.

\subsection{Feedback-delay, temporal-transfer, and adaptive analyses}
\label{subsec:eaf_exp_robustness}

Historical-feedback sensitivity was evaluated by reconstructing target-specific history features under different measurement-release delays. The temperature experiments considered delays of 60, 120, 240, and 1440 minutes, while the composition experiments considered 60, 240, 1440, and 4320 minutes. For each scenario, models were refitted using release-eligible training observations and the previously selected \textsc{History} hyperparameters. These experiments quantify how changes in feedback availability influence prediction accuracy under the prespecified release-delay scenarios introduced in Section~\ref{subsec:eaf_cutoff}.

A separate retrospective temporal-transfer analysis evaluated the selected history-informed models on the final chronological 20\% of each prediction cohort. Each model was refitted using eligible development-period observations and the hyperparameters selected in the third development fold, without further optimization on the later period. During sequential evaluation, model parameters remained fixed, while historical-outcome descriptors were updated as earlier measurements became eligible. The later period had also been examined during preceding research iterations and is therefore treated as a retrospective evaluation segment.

Adaptive residual correction was calibrated using the first half of the final development fold's chronological out-of-fold predictions. The candidate configurations included correction strengths $\lambda\in\{0,0.25,0.5,1\}$, memory lengths $L\in\{5,20,50\}$, and uniform or exponentially decaying weights with a 48-hour half-life. During later-period evaluation, residual memory was initialized from the evaluation sequence and updated only when earlier prediction outcomes became eligible. This procedure enables sequential correction of a fixed history-informed regressor without repeated model training.

\subsection{Performance measures and paired uncertainty}
\label{subsec:eaf_exp_metrics}

Predictive performance was evaluated primarily using RMSE and mean absolute error (MAE). Given $n$ observed target values $y_i$ and their corresponding predictions $\widehat{y}_i$, these metrics are defined as
\begin{equation}
\begin{aligned}
\operatorname{RMSE}
&= \sqrt{\frac{1}{n}\sum_{i=1}^{n}
(y_i-\widehat{y}_i)^2},\\[4pt]
\operatorname{MAE}
&= \frac{1}{n}\sum_{i=1}^{n}
\left|y_i-\widehat{y}_i\right|.
\end{aligned}
\label{eq:eaf_s5_metrics}
\end{equation}

Temperature prediction errors are reported in $^\circ$C, whereas chemical-composition prediction errors are expressed in wt.\%. Development-period performance metrics were calculated by pooling predictions from the three non-overlapping outer chronological validation folds, rather than averaging fold-specific RMSE values. RMSE was adopted as the principal criterion for hyperparameter optimization and model comparison, while MAE was used as a complementary measure of absolute prediction error.

To quantify the predictive contribution of historical outcome information, \textsc{History} and \textsc{Temporal} predictions were paired by target, regression algorithm, outer validation fold, and heat identifier. The performance difference was defined as
\begin{equation}
\Delta_{\mathrm{RMSE}}
=
\operatorname{RMSE}_{\mathrm{Temporal}}
-
\operatorname{RMSE}_{\mathrm{History}},
\label{eq:eaf_s5_rmse_difference}
\end{equation}
where a positive $\Delta_{\mathrm{RMSE}}$ indicates improved prediction accuracy from incorporating historical feedback. Exploratory 95\% confidence intervals were estimated using 300 paired, within-fold circular moving-block bootstrap replications, with blocks of 30 consecutive heats. This procedure preserves matched \textsc{Temporal} and \textsc{History} predictions while approximately accounting for temporal dependence among neighboring heats. Additional paired uncertainty analyses were conducted for the leading temperature regressors and the later-period adaptive residual-correction experiments.

Because multiple algorithm--target comparisons were evaluated, the resulting confidence intervals were interpreted as exploratory measures of uncertainty rather than multiplicity-adjusted hypothesis tests. Similarly, differences between the lowest observed model errors were interpreted with consideration of their magnitude and uncertainty rather than as definitive evidence of universal algorithmic superiority. To complement average prediction-error measures, the proportions of temperature predictions exceeding selected absolute-error thresholds were also evaluated. These thresholds provide descriptive measures of large prediction errors and do not represent independently verified operational acceptance limits.

\subsection{Model interpretation and exploratory DL comparison}
\label{subsec:eaf_exp_interpretation}

To investigate model-specific feature contributions, SHAP was applied to target-specific diagnostic estimators constructed using recursive feature elimination with cross-validation (RFECV). For the third chronological development fold, RFECV was performed exclusively on the corresponding training observations using three inner chronological splits with release-eligibility purging. Feature selection employed the fold-specific tuned \textsc{History} estimator, negative RMSE as the scoring criterion, an elimination step of five predictors, and a minimum of five retained features. Each diagnostic model was subsequently refitted on the same training partition using its RFECV-selected predictors.

SHAP values were computed on randomly sampled observations from the subsequent fold-3 forward-validation partition, separately from the observations used for feature selection and model fitting. Tree-based estimators were analyzed using TreeExplainer, linear estimators using LinearExplainer where applicable, and other estimators using permutation-based SHAP. The explanation sample size depended on the model family, with 20 observations for Random Forest, ExtraTrees, and XGBoost, up to 80 for other applicable explainers, and 10 for permutation-based explanations. The tree explainers used the model's tree-path-dependent reference, while the sampled validation features provided the background for linear and permutation-based explanations. Mean absolute SHAP values were calculated separately for each prediction target. These analyses characterize predictive associations within the selected-feature diagnostic models, whereas the primary benchmark evaluates the complete 47-feature \textsc{History} representation.

Four DL architectures were evaluated to investigate the predictive capabilities of neural network models on the proposed heat-level representations: multilayer perceptron (MLP), LSTM, gated recurrent unit (GRU), and one-dimensional convolutional neural network (1D CNN). All architectures were assessed for tapping temperature and the twelve chemical-composition targets using chronological development-period validation. The recurrent and convolutional networks learned from sequences of consecutive heat-level observations rather than raw within-heat sensor measurements. The neural models employed compact architectures and a smaller hyperparameter optimization budget than the classical regression benchmark. This experiment provides a complementary architectural comparison to assess the effectiveness of engineered historical-feedback representations. The results are presented in Section~\ref{subsec:eaf_s6_dl}.

\begin{table}[!ht]
\caption{Summary of the experimental configurations and their corresponding evaluation roles.}
\label{tab:eaf_exp_settings}
\centering
\footnotesize
\setlength{\tabcolsep}{4pt}
\renewcommand{\arraystretch}{1.14}

\begin{tabular}{@{}p{3.8cm}p{8.1cm}@{}}
\toprule
\textbf{Experimental component} &
\textbf{Implemented configuration} \\
\midrule

Prediction tasks &
Tapping temperature (one target) and final steel composition (twelve independently modeled elemental targets). \\

Feature representations &
\textsc{Static} (19 predictors), \textsc{Temporal} (39 predictors), and \textsc{History} (47 predictors), constructed using cutoff-eligible information. \\

Chronological data partitioning &
First 80\% of heats for model development; remaining 20\% for retrospective later-period evaluation, independently partitioned for each prediction task. \\

Model validation &
Three expanding-window outer chronological folds and three inner chronological folds for hyperparameter optimization; training labels purged according to assumed outcome-release times. \\

Hyperparameter optimization &
Twelve regression algorithms; up to 24 candidate configurations per algorithm, target, feature representation, and outer fold; random seed 2026. \\

Baselines and feature comparisons &
Outer-training median and last-20-released-outcome median baselines; matched \textsc{Static}, \textsc{Temporal}, and \textsc{History} comparisons; focused historical-feature-group ablation on the third development fold. \\

Historical-feedback sensitivity &
Assumed outcome-release delays of 1, 2, 4, and 24\,h for temperature and 1, 4, 24, and 72\,h for chemical composition. \\

Retrospective adaptive evaluation &
Selected history-informed models refitted on development observations using fold-3 tuned hyperparameters; adaptive online residual correction using previously released prediction errors without model retraining. \\

Performance evaluation and uncertainty &
RMSE and MAE as primary accuracy metrics; paired moving-block bootstrap confidence intervals; relative RMSE improvements; and temperature absolute-error exceedance rates. \\

Model interpretation and additional comparisons &
Target-specific SHAP analyses for temperature and all twelve chemical elements using RFECV-selected diagnostic models; exploratory comparisons with MLP, LSTM, GRU, and 1D CNN across thirteen prediction targets. \\

\bottomrule
\end{tabular}
\end{table}

\section{Results and Discussion}
\label{sec:results}

This section evaluates the availability-aware historical-feedback framework from four complementary perspectives: comparative predictive accuracy, the information added by released outcomes, the effect of feedback timeliness, and sequential performance as new heats are processed. Temperature and each of the twelve elemental-composition outcomes are assessed under the chronological protocol of Section~\ref{sec:eaf_experiments}. The development-period benchmark reports pooled out-of-fold errors from three chronological validation folds, whereas the later-period study evaluates frozen models and sequential residual correction on the final 20\% of the heat sequence. All numerical comparisons use RMSE and MAE in the corresponding target units. The figures complement the performance tables by illustrating cross-algorithm patterns, feedback sensitivity, time-local prediction errors, and feature attribution, providing additional insights into model behavior and predictive performance.

\subsection{Development-period prediction performance}
\label{subsec:eaf_s6_benchmark}

The first experiment evaluates how prediction accuracy changes across three progressively enriched feature representations: \textsc{Static}, containing current-heat operational descriptors; \textsc{Temporal}, incorporating additional within-heat temporal information; and \textsc{History}, introducing previously released outcome measurements. Table~\ref{tab:eaf_s6_temp} compares these representations across twelve regression algorithms under chronological validation, allowing the predictive contribution of each information level to be examined within the same model family.

For tapping-temperature prediction, the \textsc{History} configuration achieves lower RMSE than \textsc{Temporal} across all twelve regressors. ExtraTrees provides the lowest observed RMSE of 12.290~$^\circ$C, compared with 13.011~$^\circ$C using temporal features alone, representing a 5.54\% reduction. Random Forest and Gradient Boosting achieve similarly competitive results of 12.295 and 12.322~$^\circ$C, respectively. The consistent improvements across different algorithms highlight the predictive value of released historical outcomes beyond current-heat measurements. Furthermore, the history-informed ExtraTrees model outperforms the last-20-released-outcome median baseline defined in Section~\ref{subsec:eaf_exp_baselines}, reducing RMSE from 12.769 to 12.290~$^\circ$C (3.75\%). This demonstrates the benefit of combining historical outcome statistics with current-heat transformer activity, charging characteristics, and gas and oxygen-flow information rather than relying solely on recent temperature measurements.

\begin{table}[!ht]
\centering
\scriptsize
\setlength{\tabcolsep}{3pt}
\renewcommand{\arraystretch}{1.15}
\caption{Development-period performance of twelve temperature regressors under chronological validation. Reported errors ($^\circ$C) are calculated from pooled outer-fold predictions.}
\label{tab:eaf_s6_temp}
\begin{tabular}{@{}lrrrr@{}}
\toprule
\textbf{Algorithm} & \textbf{STATIC RMSE} & \textbf{TEMPORAL RMSE} & \textbf{HISTORY RMSE} & \textbf{HISTORY MAE} \\
\midrule
ExtraTrees & 13.007 & 13.011 & 12.290 & 9.130 \\
Random Forest & 13.103 & 13.079 & 12.295 & 9.186 \\
Gradient Boosting & 13.091 & 13.028 & 12.322 & 9.161 \\
ElasticNet & 13.113 & 13.025 & 12.340 & 9.135 \\
Lasso & 13.106 & 13.012 & 12.340 & 9.135 \\
AdaBoost & 13.301 & 13.272 & 12.349 & 9.245 \\
Bayesian Ridge & 13.038 & 12.954 & 12.357 & 9.107 \\
SVR & 12.945 & 13.007 & 12.357 & 9.169 \\
XGBoost & 12.996 & 12.929 & 12.365 & 9.245 \\
Ridge & 13.054 & 13.048 & 12.465 & 9.163 \\
KNN & 13.351 & 13.288 & 12.722 & 9.498 \\
Decision Tree & 14.081 & 13.750 & 12.878 & 9.590 \\
\bottomrule
\end{tabular}
\end{table}

Table~\ref{tab:eaf_s6_chem} presents the corresponding chemical-composition results, showing that the best-performing \textsc{History} regressor varies across elemental targets. Random Forest achieves the lowest RMSE for carbon, phosphorus, molybdenum, and nitrogen; ExtraTrees for silicon, copper, nickel, and tin; XGBoost for manganese; Ridge for sulfur; and Lasso for chromium and arsenic. These differences highlight the importance of target-specific model selection, as individual elements exhibit distinct chemical characteristics and predictive relationships. The proposed framework accommodates this diversity through a consistent historical-feedback representation applicable across multiple regression families.

The largest relative RMSE reductions are observed for molybdenum (18.80\%) and phosphorus (17.10\%) using Random Forest, followed by sulfur (12.92\%) using Ridge, and copper (12.56\%) and nickel (11.66\%) using ExtraTrees. These reductions compare the lowest \textsc{History} and \textsc{Temporal} RMSE values for each element, with model selection performed independently for the two configurations. Smaller improvements are observed for silicon (0.31\%) and nitrogen (1.78\%), highlighting the target-dependent benefits of historical feedback. Nitrogen has fewer development-validation observations ($n=1{,}386$) due to its lower and temporally varying assay coverage, as discussed in Section~\ref{subsec:eaf_cohorts}. Consequently, nitrogen prediction performance is assessed on heats with observed nitrogen labels, and its interpretation accounts for differences in assay availability across chronological periods. Overall, these findings demonstrate that previously released chemistry measurements provide valuable complementary information for predicting final steel composition, emphasizing the effectiveness of combining historical feedback with target-specific regression models.

\begin{table}[!ht]
\centering
\scriptsize
\setlength{\tabcolsep}{3pt}
\renewcommand{\arraystretch}{1.15}
\caption{Element-wise development-period results for the lowest observed HISTORY RMSE among twelve tuned regressors. Relative reduction is computed against the lowest TEMPORAL RMSE for the same target. All errors are in wt.\%; model selections are made independently for each element and feature configuration.}
\label{tab:eaf_s6_chem}
\begin{tabular}{@{}llrrrr@{}}
\toprule
\textbf{Element} & \textbf{HISTORY model} & \textbf{$n$} & \textbf{RMSE (wt.\%)} & \textbf{MAE (wt.\%)} & \textbf{RMSE reduction (\%)} \\
\midrule
C & RandomForest & 2004 & 0.02464 & 0.01550 & 2.65 \\
Si & ExtraTrees & 2004 & 0.007579 & 0.002867 & 0.31 \\
Mn & XGBoost & 2004 & 0.03403 & 0.02203 & 7.06 \\
P & RandomForest & 2003 & 0.002893 & 0.001837 & 17.10 \\
S & Ridge & 2004 & 0.009383 & 0.006714 & 12.92 \\
Cu & ExtraTrees & 2004 & 0.02320 & 0.01717 & 12.56 \\
Cr & Lasso & 2004 & 0.04222 & 0.02274 & 8.64 \\
Mo & RandomForest & 2004 & 0.02917 & 0.01335 & 18.80 \\
Ni & ExtraTrees & 2004 & 0.02283 & 0.01586 & 11.66 \\
As & Lasso & 2004 & 0.000748 & 0.000490 & 11.07 \\
Sn & ExtraTrees & 2004 & 0.001977 & 0.001174 & 7.04 \\
N & RandomForest & 1386 & 0.01487 & 0.005037 & 1.78 \\
\bottomrule
\end{tabular}
\end{table}

\subsection{Contribution of historical outcomes across algorithm families}
\label{subsec:eaf_s6_history}

A central question is whether the improvement associated with \textsc{History} features depends on choosing a particular regression algorithm. We address this through matched \textsc{Temporal}-versus-\textsc{History} comparisons for every target and every model family. For each target and regression family, the two configurations share the same outer-fold boundaries and validation observations. Their feature representations differ through the inclusion of previously released outcomes, with hyperparameters optimized separately under the same chronological validation protocol. Thus, a positive relative RMSE reduction measures the incremental benefit of historical feedback within a fixed learning approach, rather than an advantage obtained by switching to a more flexible algorithm.

\begin{figure}[!ht]
\centering
\includegraphics[width=\linewidth]{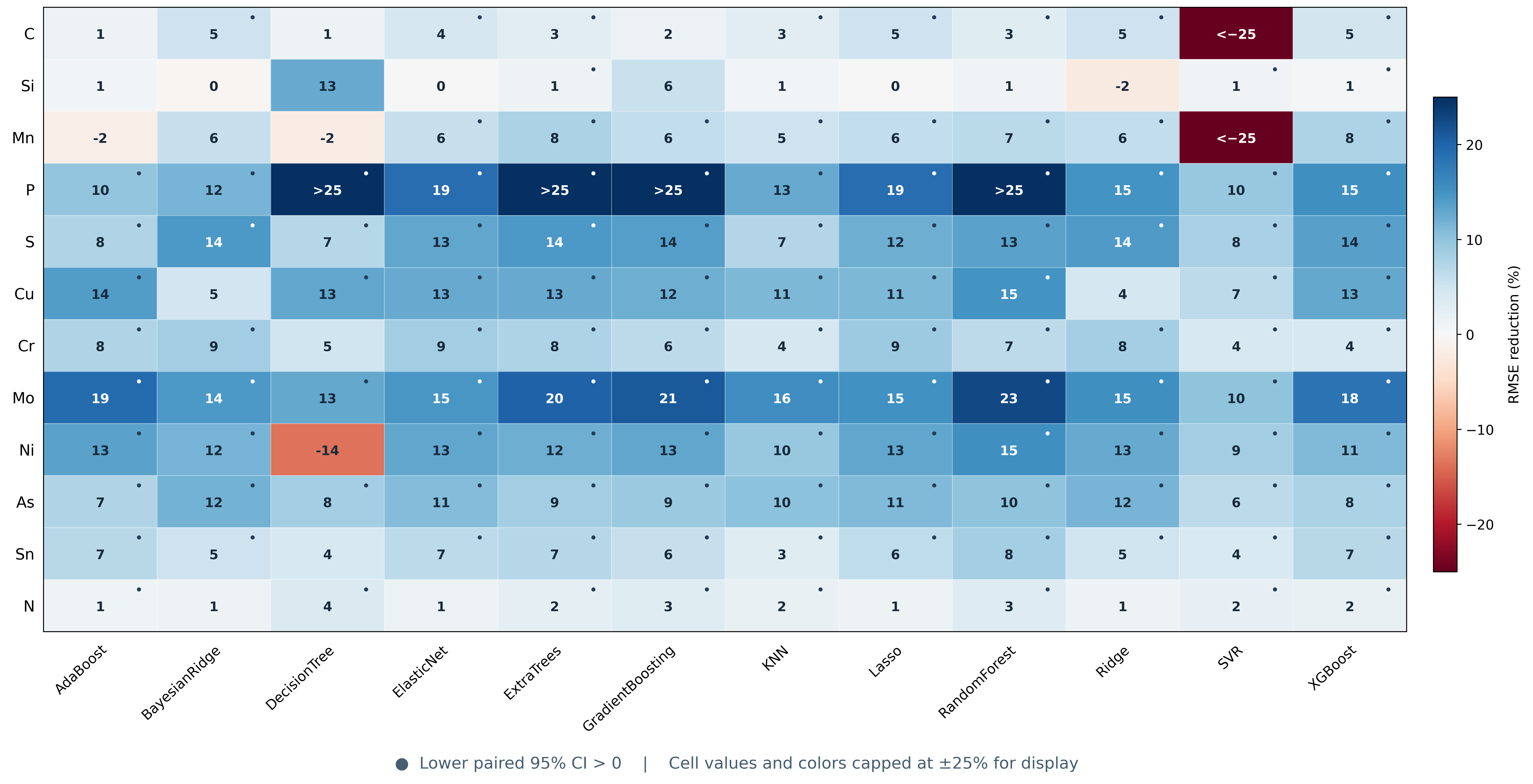}
\caption{Element-wise effect of released historical outcomes, comparing HISTORY and TEMPORAL predictors within the same algorithm. Cells show percentage reduction in pooled RMSE; dots indicate exploratory paired 95\% intervals above zero. Colors are capped at $\pm25\%$ for legibility, without truncating the underlying data.}
\label{fig:eaf_s6_history}
\end{figure}

The result is strikingly consistent across the benchmark: 147 of the 156 algorithm--target comparisons favor the history-informed representation. All twelve temperature regressors benefit, as do 135 of the 144 chemistry combinations. Figure~\ref{fig:eaf_s6_history} resolves this aggregate finding into an element-by-algorithm heatmap. The broad positive pattern extends across linear regularization, tree ensembles, boosting, instance-based regression, and kernel methods. This breadth supports a model-independent interpretation: previously released outcomes carry information that several different estimators can exploit, even when their internal learning mechanisms are quite different.

The heatmap also reveals meaningful differences among chemical elements. Phosphorus and molybdenum show particularly strong reductions across multiple algorithms, with substantial gains also visible for sulfur, copper, nickel, and arsenic. Silicon and nitrogen tend to exhibit comparatively modest changes. These differences suggest that the benefit of outcome history is linked to target-specific behavior rather than simply to increasing the number of input variables. An element with persistent heat-to-heat patterns may benefit strongly from its recent assay history; a target whose variation is less predictable from released results can exhibit smaller gains. This interpretation is consistent with the target-dependent SHAP patterns discussed in Section~\ref{subsec:eaf_s6_shap}.

The paired moving-block bootstrap adds an uncertainty perspective to these observed improvements. Positive lower confidence bounds occur in 130 of the 156 comparisons, indicating that the favorable direction persists under resampling for a large majority of algorithm--target combinations. The intervals are reported as exploratory uncertainty summaries for the collection of matched comparisons. Figure~\ref{fig:eaf_s6_history} also retains the unfavorable individual cases, including pronounced prediction errors in some polynomial-kernel SVR configurations. The display range is capped at $\pm25\%$ to make the overall pattern legible. Collectively, the findings identify historical-outcome availability as a broadly useful source of predictive information, while the variation across elements provides a basis for selecting how the representation should be used in different chemistry tasks.

\subsection{Historical-feature ablation and assumed release-delay sensitivity}
\label{subsec:eaf_s6_ablation}

\begin{figure}[!ht]
\centering
\includegraphics[width=\linewidth]{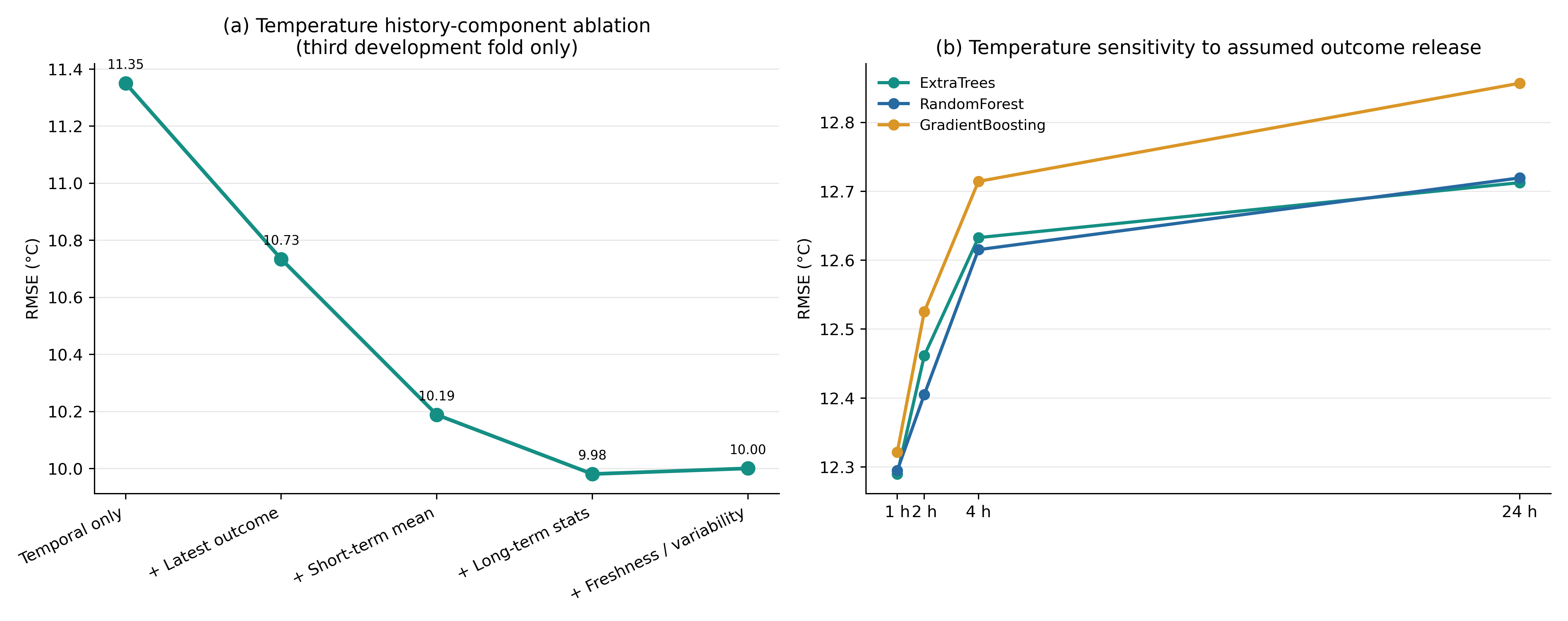}
\caption{Historical-feedback analysis for EAF temperature prediction: (a) incremental historical-feature ablation using ExtraTrees on the third chronological development fold; (b) pooled temperature RMSE for ExtraTrees, Random Forest, and Gradient Boosting under assumed outcome-release delays of 1, 2, 4, and 24 hours.}
\label{fig:eaf_s6_ablation}
\end{figure}

The overall \textsc{History} improvement raises two further questions: which historical descriptors contribute most to prediction accuracy, and how does their value change when outcome measurements become available later? Figure~\ref{fig:eaf_s6_ablation} addresses these questions for temperature prediction through staged historical-feature ablation and feedback-release sensitivity analysis. The corresponding chemical-composition experiments are reported in Tables~\ref{tab:eaf_s6_chem_ablation} and~\ref{tab:eaf_s6_chem_delay}, allowing the contributions of historical information and its availability to be examined across both prediction tasks.

Figure~\ref{fig:eaf_s6_ablation}(a) presents the ExtraTrees temperature ablation on the third chronological development fold. Starting with the \textsc{Temporal} representation, RMSE is 11.35~$^\circ$C. Adding the latest released temperature reduces RMSE to 10.73~$^\circ$C, representing the largest individual improvement. Incorporating the mean of the five most recently released temperatures further reduces RMSE to 10.19~$^\circ$C, while adding longer-term historical statistics achieves the lowest observed RMSE of 9.98~$^\circ$C. The complete eight-descriptor representation produces a comparable RMSE of 10.00~$^\circ$C. These results demonstrate that recent outcomes and their moving averages capture a substantial portion of the predictive information provided by historical feedback.

The ablation also clarifies the roles of different historical descriptors. The latest released temperature reflects recent furnace behavior, short-term averages capture prevailing operating trends while reducing sensitivity to individual observations, and longer-term statistics provide information about historical variability. Additional freshness, availability, and trend descriptors further characterize the historical information accessible at prediction time. The analysis therefore identifies the relative contributions of historical-feature groups within the proposed framework.

Figure~\ref{fig:eaf_s6_ablation}(b) examines the sensitivity of ExtraTrees, Random Forest, and Gradient Boosting to assumed outcome-release delays of 1, 2, 4, and 24 hours, using pooled chronological development-validation results. All three regressors exhibit increasing RMSE as historical measurements become available later, with a particularly noticeable deterioration between the one-hour and four-hour scenarios. This consistent pattern indicates that the predictive contribution of historical feedback depends not only on the information contained in previous outcomes but also on how recently those outcomes were measured and released.

Together, these experiments establish the importance of both historical-feature composition and feedback timeliness for EAF temperature prediction. The findings show that recent outcome measurements and their statistical summaries provide informative complements to current-heat operational data, while timely access to these measurements strengthens their predictive utility. This highlights the relevance of incorporating measurement-availability considerations into data-driven EAF prediction frameworks.

The historical-feature ablation was also evaluated for four representative chemical-composition targets: phosphorus (P), molybdenum (Mo), silicon (Si), and nitrogen (N). Table~\ref{tab:eaf_s6_chem_ablation} reports the resulting RMSE values for the five cumulative feature configurations on the third chronological development-validation fold. Random Forest was used for P, Mo, and N, while ExtraTrees was used for Si, with the previously selected fold-specific \textsc{History} hyperparameters retained across configurations.

\begin{table}[!ht]
\centering
\scriptsize
\setlength{\tabcolsep}{5pt}
\renewcommand{\arraystretch}{1.15}
\caption{Historical-feature-group ablation for four chemical-composition targets on the third chronological development-validation fold. RMSE values are in wt.\%. A0 uses 39 temporal predictors; A1--A3 add historical-feature groups cumulatively; A4 uses the complete 47-feature \textsc{History} representation.}
\label{tab:eaf_s6_chem_ablation}
\begin{tabular}{llccccc}
\toprule
Target & Model & A0 & A1 & A2 & A3 & A4 \\
\midrule
P  & Random Forest & 0.004097 & 0.003802 & 0.003764 & 0.003766 & \textbf{0.003716} \\
Mo & Random Forest & 0.038747 & 0.027899 & 0.027888 & 0.028437 & \textbf{0.026694} \\
Si & ExtraTrees    & 0.010743 & 0.010719 & 0.010695 & 0.010677 & \textbf{0.010585} \\
N  & Random Forest & 0.029175 & 0.028890 & 0.028847 & 0.028800 & \textbf{0.028733} \\
\bottomrule
\end{tabular}
\end{table}

The complete historical-feedback representation (A4) achieves the lowest RMSE among the five evaluated configurations for all four elements. Relative to the temporal-only configuration (A0), RMSE decreases by 31.11\% for Mo and 9.28\% for P, while smaller reductions of 1.47\% and 1.51\% are observed for Si and N, respectively. For Mo, incorporating the latest released measurement (A1) provides the largest individual improvement, whereas subsequent historical summaries contribute additional predictive information. The different improvement patterns across elements indicate that the importance of individual historical-feedback components is target-dependent. These findings complement the temperature ablation in Figure~\ref{fig:eaf_s6_ablation}(a) and support the use of a common historical-feature construction with independently fitted, target-specific regression models.

The influence of assumed outcome-release delay was further evaluated for all twelve chemical-composition targets. Table~\ref{tab:eaf_s6_chem_delay} summarizes the pooled chronological development-validation RMSE under release-delay scenarios of 1, 4, 24, and 72 hours. For each element, the selected \textsc{History} regression algorithm was retained and the historical descriptors were reconstructed according to the corresponding release-availability scenario.

\begin{table}[!ht]
\centering
\scriptsize
\setlength{\tabcolsep}{7pt}
\renewcommand{\arraystretch}{1.12}
\caption{Chemical-composition prediction RMSE under different assumed historical-outcome release delays. Results are pooled across the three chronological development-validation folds and reported in wt.\%.}
\label{tab:eaf_s6_chem_delay}
\begin{tabular}{lcccc}
\toprule
Element & 1 h & 4 h & 24 h & 72 h \\
\midrule
C  & 0.024640 & 0.024860 & 0.024931 & 0.025248 \\
Si & 0.007579 & 0.007572 & 0.007620 & 0.007639 \\
Mn & 0.034028 & 0.035049 & 0.035616 & 0.036105 \\
P  & 0.002893 & 0.003055 & 0.003101 & 0.003107 \\
S  & 0.009383 & 0.010622 & 0.011749 & 0.010635 \\
Cu & 0.023199 & 0.024782 & 0.025209 & 0.025616 \\
Cr & 0.042217 & 0.045757 & 0.046108 & 0.046272 \\
Mo & 0.029166 & 0.034973 & 0.038274 & 0.039949 \\
Ni & 0.022831 & 0.024374 & 0.025556 & 0.025394 \\
As & 0.000748 & 0.000808 & 0.000825 & 0.000835 \\
Sn & 0.001977 & 0.002070 & 0.002055 & 0.002065 \\
N  & 0.014866 & 0.014940 & 0.015083 & 0.015224 \\
\bottomrule
\end{tabular}
\end{table}

The sensitivity analysis shows that the assumed timeliness of historical measurements influences chemical-composition prediction accuracy. All twelve elements exhibit higher RMSE under the 72-hour delay than under the baseline one-hour scenario, although individual error trends are not strictly monotonic across intermediate delays. Molybdenum exhibits the largest relative deterioration, with RMSE increasing from 0.029166 to 0.039949 wt.\% (36.97\%). Phosphorus RMSE increases by 7.40\%, while silicon and nitrogen exhibit smaller changes of approximately 0.79\% and 2.41\%, respectively. These findings demonstrate that the predictive value of historical chemical measurements depends on both the elemental target and the timeliness of feedback availability.

Together, the temperature and chemistry results show that historical outcomes provide valuable predictive information beyond current-heat measurements, while the magnitude of improvement depends on the target, composition of the historical feature set, and assumed outcome-release delay.

\subsection{Retrospective later-period evaluation and adaptive correction}
\label{subsec:eaf_s6_adaptive}

To assess sequential prediction performance, the selected \textsc{History} models are refitted on eligible development observations and evaluated on the final chronological 20\% of heats. During this evaluation, the fitted regressors remain unchanged, while historical descriptors are updated as earlier outcomes become available. The adaptive residual-feedback mechanism introduced in Section~\ref{subsec:eaf_framework_adaptive} further adjusts predictions using previously released prediction errors. Unlike the eight historical features, which summarize observed outcomes, this correction captures recent discrepancies between model predictions and actual measurements, enabling sequential adjustment without model retraining.

Table~\ref{tab:eaf_s6_later} compares frozen and adaptively corrected predictions across all thirteen targets. For temperature, adaptive correction reduces the ExtraTrees RMSE from 12.502 to 12.275~$^\circ$C, corresponding to an improvement of 0.227~$^\circ$C (1.82\%). MAE similarly decreases from 9.755 to 9.608~$^\circ$C, while the paired bootstrap interval for the RMSE reduction is $[0.148,0.314]~^\circ$C. These improvements demonstrate the ability of recently observed prediction errors to refine subsequent temperature estimates.

For chemical composition, the adaptive response varies across elements. Nitrogen shows the largest relative RMSE reduction (2.9\%), followed by manganese (1.6\%) and phosphorus (1.3\%), with smaller improvements for silicon, chromium, nickel, and tin. The selected correction strength is zero for carbon and sulfur, while copper, molybdenum, and arsenic exhibit slight increases in RMSE. These results highlight the target-specific nature of residual feedback: adaptation is most beneficial when recent prediction errors contain persistent information relevant to subsequent heats. The target-wise calibration of correction strength enables the framework to accommodate these differences.

\begin{figure}[!ht]
\centering
\includegraphics[width=\linewidth]{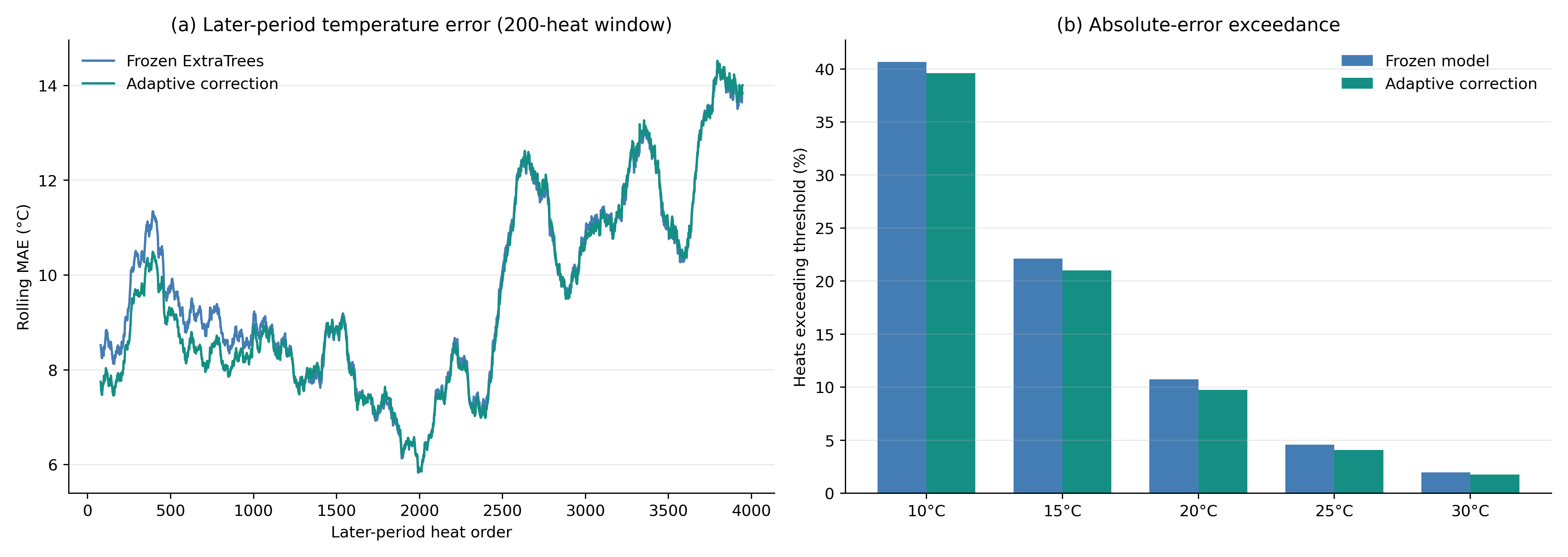}
\caption{Later-period temperature diagnostics showing (a) rolling MAE across successive heats and (b) the proportion of absolute prediction errors exceeding illustrative temperature thresholds, for frozen and adaptively corrected predictions.}
\label{fig:eaf_s6_adaptive}
\end{figure}

Figure~\ref{fig:eaf_s6_adaptive} further examines temperature prediction reliability throughout the later-period sequence. Panel (a) presents rolling MAE, revealing changes in prediction accuracy across successive heats and periods where adaptive correction reduces local errors. Panel (b) compares the proportions of predictions exceeding selected absolute-error thresholds, showing lower exceedance frequencies after correction. Together, these analyses demonstrate that adaptive residual feedback can improve both aggregate accuracy and the frequency of larger prediction errors while preserving the computational efficiency of a fixed regression model.

\begin{table}[!ht]
\centering
\scriptsize
\setlength{\tabcolsep}{3pt}
\renewcommand{\arraystretch}{1.14}
\caption{Retrospective later-period performance for frozen predictions and adaptive residual-corrected predictions. RMSE and MAE are reported in $^\circ$C for temperature and wt.\% for elemental composition; positive RMSE change indicates improvement. The evaluation uses the final chronological 20\% of heats.}
\label{tab:eaf_s6_later}
\begin{tabular}{@{}llrrrrrr@{}}
\toprule
\textbf{Target} & \textbf{Model} & \textbf{$n$} & \textbf{Frozen RMSE} & \textbf{Adaptive RMSE} & \textbf{Frozen MAE} & \textbf{Adaptive MAE} & \textbf{RMSE gain (\%)} \\
\midrule
Temperature & ExtraTrees & 3948 & 12.502 & 12.275 & 9.755 & 9.608 & +1.8 \\
C & RandomForest & 668 & 0.02496 & 0.02496 & 0.01551 & 0.01551 & +0.0 \\
Si & ExtraTrees & 669 & 0.004852 & 0.004843 & 0.003456 & 0.003331 & +0.2 \\
Mn & XGBoost & 669 & 0.03122 & 0.03072 & 0.02016 & 0.01928 & +1.6 \\
P & RandomForest & 669 & 0.002880 & 0.002843 & 0.002205 & 0.002146 & +1.3 \\
S & Ridge & 669 & 0.009620 & 0.009620 & 0.006915 & 0.006915 & +0.0 \\
Cu & ExtraTrees & 669 & 0.02047 & 0.02074 & 0.01431 & 0.01470 & -1.3 \\
Cr & Lasso & 669 & 0.02884 & 0.02879 & 0.01890 & 0.01837 & +0.2 \\
Mo & RandomForest & 669 & 0.02545 & 0.02552 & 0.01030 & 0.01034 & -0.3 \\
Ni & ExtraTrees & 669 & 0.01299 & 0.01297 & 0.00971 & 0.00963 & +0.2 \\
As & Lasso & 669 & 0.001200 & 0.001201 & 0.000594 & 0.000594 & -0.1 \\
Sn & ExtraTrees & 668 & 0.001284 & 0.001281 & 0.000974 & 0.000977 & +0.2 \\
N & RandomForest & 382 & 0.009890 & 0.009605 & 0.006790 & 0.006284 & +2.9 \\
\bottomrule
\end{tabular}
\end{table}

\subsection{Exploratory comparison with DL models}
\label{subsec:eaf_s6_dl}

Four DL architectures were evaluated for all thirteen prediction targets: MLP, LSTM, GRU, and 1D CNN. This analysis investigates whether more complex neural architectures offer additional predictive advantages over classical regressors equipped with engineered historical-outcome features.

Table~\ref{tab:eaf_s6_dl} summarizes the DL prediction results. For temperature, GRU achieves the lowest RMSE among neural network models of 12.470~$^\circ$C, compared with 12.290~$^\circ$C for the history-informed ExtraTrees model reported in Table~\ref{tab:eaf_s6_temp}. Across all thirteen targets, the classical \textsc{History} benchmarks achieve lower RMSE in 51 of 52 neural-model--target comparisons. The only exception is nitrogen, for which 1D CNN achieves 0.01484~wt.\%, compared with 0.01487~wt.\% for the classical reference. These results highlight the strong predictive performance of historical-feedback-enhanced classical regressors across both temperature and elemental-composition targets.

\begin{table}[!ht]
\centering
\caption{Development-period RMSE of four DL models for EAF temperature and chemical composition prediction. Temperature errors are in $^\circ$C and composition errors in wt.\%.}
\label{tab:eaf_s6_dl}

\begingroup
\scriptsize
\setlength{\tabcolsep}{8pt}
\renewcommand{\arraystretch}{1.05}

\begin{tabular}{@{}lrrrr@{}}
\toprule
\textbf{Target} &
\textbf{MLP} &
\textbf{LSTM} &
\textbf{GRU} &
\textbf{1D CNN} \\
\midrule
Temp. & 12.589 & 12.957 & 12.470 & 12.682 \\
C  & 0.02855  & 0.02721  & 0.02657  & 0.02775 \\
Si & 0.007704 & 0.007727 & 0.007783 & 0.007638 \\
Mn & 0.04122  & 0.04015  & 0.03822  & 0.03802 \\
P  & 0.003191 & 0.003057 & 0.003188 & 0.003056 \\
S  & 0.01058  & 0.01111  & 0.01057  & 0.01016 \\
Cu & 0.02434  & 0.02945  & 0.02397  & 0.02424 \\
Cr & 0.04691  & 0.04686  & 0.04611  & 0.04502 \\
Mo & 0.03498  & 0.03392  & 0.03277  & 0.03272 \\
Ni & 0.02574  & 0.02576  & 0.02489  & 0.02537 \\
As & 0.000817 & 0.000860 & 0.000795 & 0.000818 \\
Sn & 0.002049 & 0.002269 & 0.002047 & 0.002052 \\
N  & 0.01488  & 0.01492  & 0.01497  & 0.01484 \\
\bottomrule
\end{tabular}
\endgroup
\end{table}

The findings emphasize the importance of informative feature representations in EAF prediction. The proposed \textsc{History} configuration explicitly incorporates recent outcomes, moving historical averages, variability, and feedback recency, allowing classical regressors to capture relevant process patterns directly. In contrast, the LSTM, GRU, and 1D CNN models learn from sequences of consecutive heat-level records rather than high-frequency within-heat sensor measurements. Overall, the comparison demonstrates that carefully engineered historical-feedback features can provide competitive predictive accuracy without requiring complex neural architectures, while motivating further investigation of DL with richer process-level time-series data.

\begin{figure}[!ht]
\centering
\includegraphics[width=0.65\linewidth]{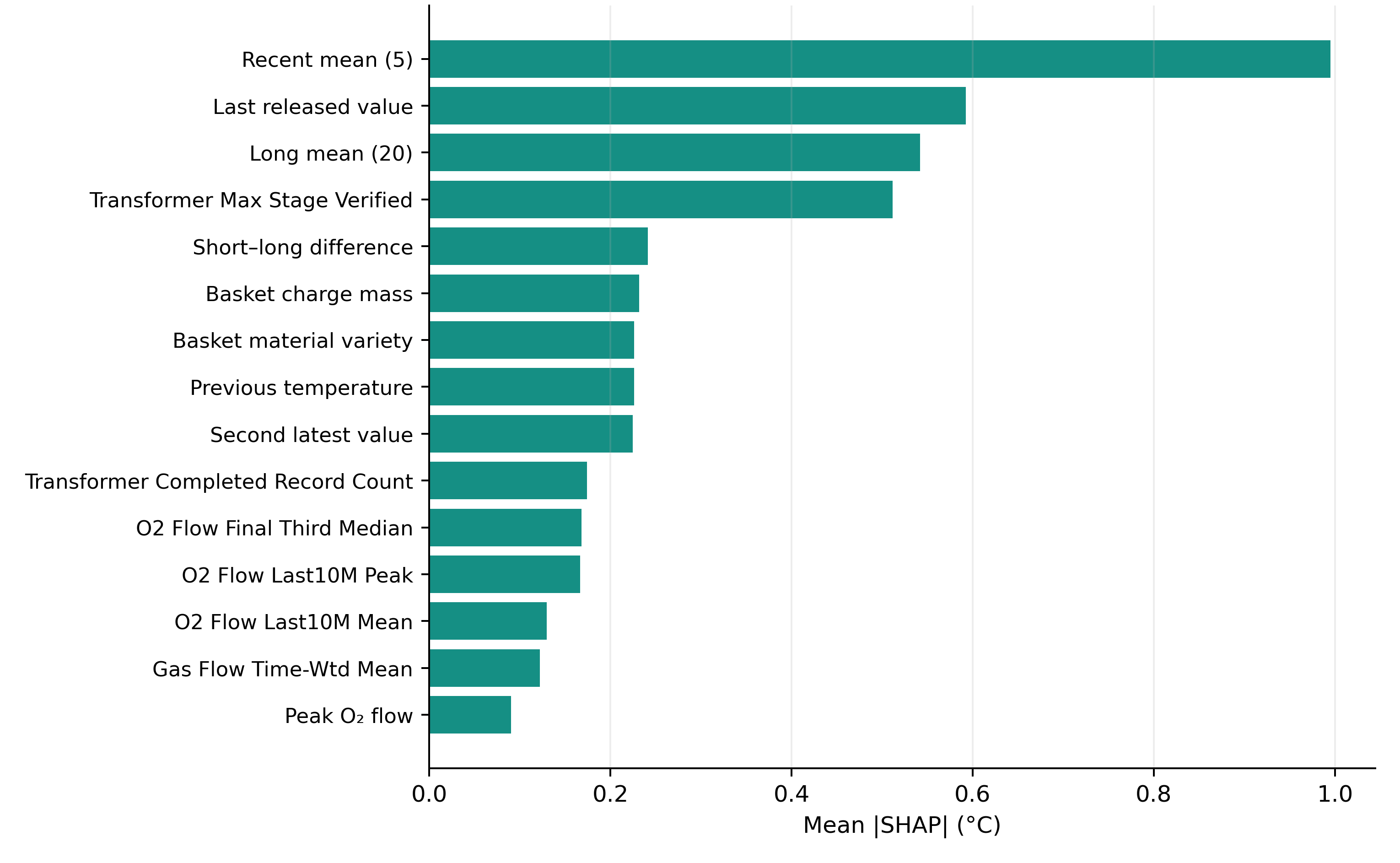}
\caption{Temperature feature-attribution diagnostics for a fold-3 RFECV-selected ExtraTrees model. Bars show mean absolute SHAP values in $^\circ$C; the displayed attributions pertain to the selected-feature diagnostic estimator.}
\label{fig:eaf_s6_shap_temp}
\end{figure}

\subsection{Target-specific feature attribution}
\label{subsec:eaf_s6_shap}

We employ SHAP-based feature attribution to examine the relative contributions of operational measurements and historical feedback to temperature and chemical-composition prediction. The analysis uses RFECV-selected diagnostic models trained on the third chronological development fold's training observations, with explanations computed on sampled forward-validation heats, as described in Section~\ref{subsec:eaf_exp_interpretation}. By analyzing each prediction target separately, the resulting attributions identify the operational and historical predictors most influential to the fitted diagnostic models.

Figure~\ref{fig:eaf_s6_shap_temp} presents the temperature feature-attribution results for the selected-feature ExtraTrees model. The mean of the five most recently released temperatures exhibits the highest mean absolute SHAP value, followed by the latest released temperature, the longer-term historical mean, and transformer-stage information. Other influential predictors include differences between short- and long-term historical averages, charging characteristics, earlier pre-cutoff temperature measurements, and oxygen-flow descriptors. These findings complement the historical-feature ablation results in Section~\ref{subsec:eaf_s6_ablation}, demonstrating that recent temperature feedback provides substantial predictive information while current-heat operational variables contribute additional context.

\begin{figure}[!ht]
\centering
\includegraphics[width=\linewidth]{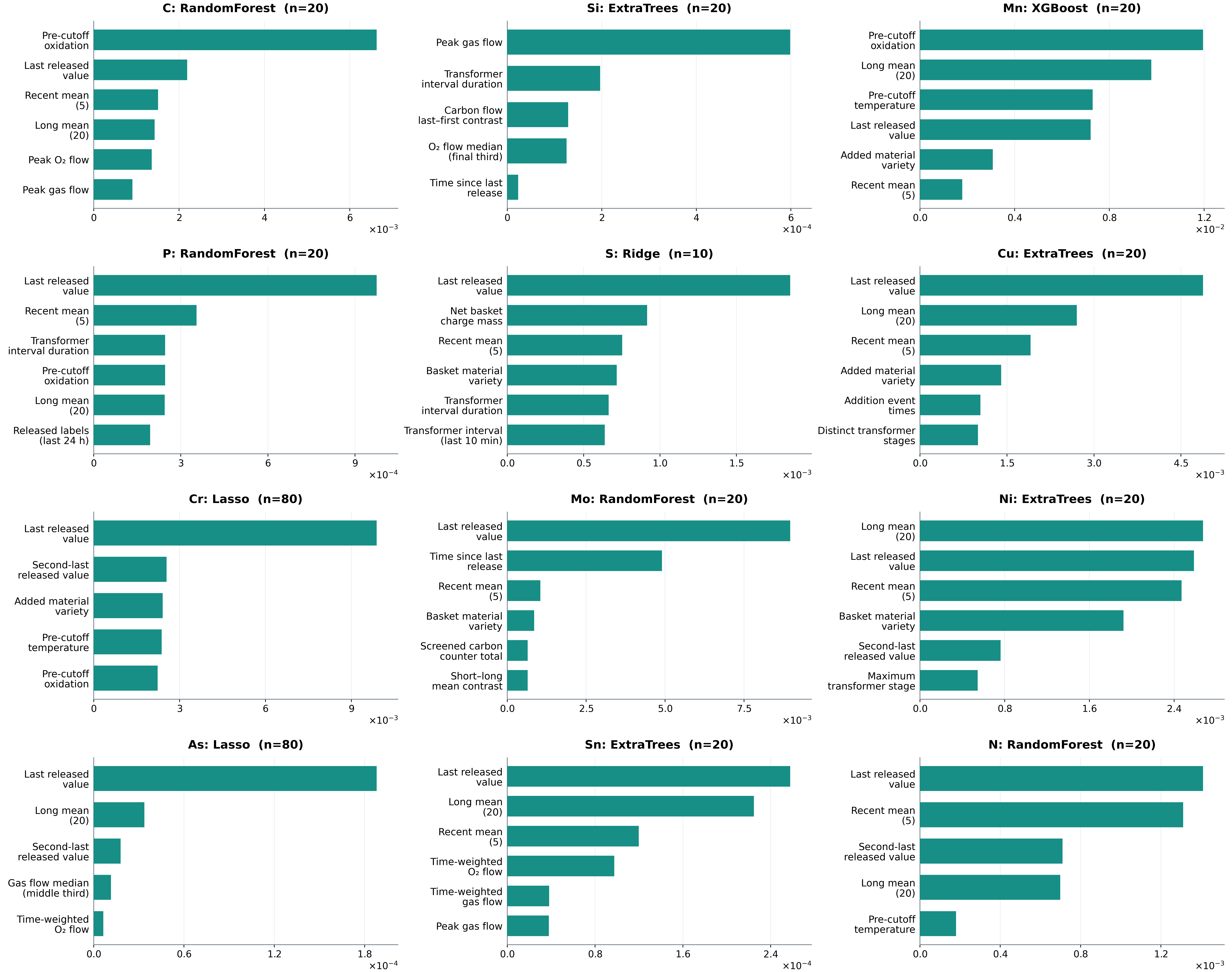}
\caption{Leading mean absolute SHAP attributions for the twelve chemistry models fitted on RFECV-selected features in development fold 3. Each panel uses the corresponding target scale; model fits and feature subsets are those of the diagnostic attribution analysis. The notation $n$ in each panel denotes the number of observations explained by SHAP.}
\label{fig:eaf_s6_shap_chem}
\end{figure}

Figure~\ref{fig:eaf_s6_shap_chem} presents the corresponding SHAP results for all twelve chemical elements. Historical outcome information is prominent for several targets, particularly phosphorus, sulfur, copper, chromium, molybdenum, arsenic, tin, and nitrogen, where the latest released elemental measurement is among the most influential predictors. Longer-term historical averages contribute strongly to manganese and nickel prediction. In contrast, carbon prediction emphasizes pre-cutoff oxidation measurements alongside historical carbon information, while silicon relies more heavily on gas-flow, transformer, and carbon-injection descriptors. The relatively small historical-feedback improvement for silicon in Table~\ref{tab:eaf_s6_chem} is consistent with this operationally dominated attribution pattern. These differences highlight how the proposed framework captures target-specific relationships between current-heat conditions and previously released chemical measurements.

Overall, the SHAP findings provide complementary insight into the predictive information used by the fitted models. Historical outcomes and their statistical summaries contribute prominently to several temperature and chemistry predictions, while operational measurements remain influential in target-dependent ways. The attributions are interpreted within each target's physical scale and describe relationships learned by the diagnostic estimators. Together with the performance and sensitivity analyses, these findings reinforce the importance of integrating historical feedback with current-process information for EAF endpoint prediction.


\section{Conclusion}
\label{sec:eaf_conclusion}

This study developed an availability-aware historical-feedback learning framework for predicting tapping temperature and the final concentrations of twelve chemical elements in EAF steelmaking. A comprehensive data-engineering pipeline integrated eight heterogeneous industrial sources through timestamp alignment, cutoff-aware preprocessing, quality screening, and process-specific feature construction. Three progressively enriched representations---\textsc{Static}, \textsc{Temporal}, and \textsc{History}---were developed to distinguish current-heat operational information, within-heat temporal behavior, and previously released outcome measurements. By explicitly incorporating historical-feedback availability into the prediction process, the framework extends conventional heat-level modeling toward a unified approach for sequential EAF endpoint prediction.

The experimental findings demonstrate the predictive value of historical outcome information across diverse regression algorithms and prediction targets. Under chronological forward validation, the \textsc{History} representation reduced RMSE in 147 of 156 matched algorithm--target comparisons, including all twelve temperature regressors. ExtraTrees achieved the lowest observed temperature RMSE of 12.290~$^\circ$C, representing a 5.54\% reduction relative to its \textsc{Temporal} counterpart. Chemical-composition prediction also benefited from historical feedback, with particularly notable relative RMSE reductions for molybdenum (18.80\%) and phosphorus (17.10\%). The variation in leading regression algorithms across chemical elements further highlights the importance of combining a consistent information representation with target-specific predictive models.

The additional analyses provide further insights into the mechanisms and effectiveness of historical feedback. Feature-group ablation identified recent released outcomes and their moving statistical summaries as important sources of predictive information, while release-delay sensitivity experiments demonstrated the influence of feedback timeliness on prediction accuracy. In the retrospective later-period evaluation, adaptive residual correction reduced temperature RMSE from 12.502 to 12.275~$^\circ$C (1.82\%) without retraining the underlying regression model, with element-dependent responses for chemical composition. Comparisons with four DL architectures demonstrated the competitiveness of classical regressors equipped with engineered historical features, while target-specific SHAP analyses highlighted the complementary predictive contributions of historical outcomes and current-heat operational measurements.

Overall, the study establishes a systematic and reproducible methodology that integrates cutoff-aware industrial data processing, historical outcome availability, chronological model evaluation, and sequential residual feedback for EAF endpoint prediction. The findings demonstrate that improving the information available to predictive models can be as important as increasing model complexity. The proposed framework provides a foundation for adaptive, data-driven decision support in steelmaking, with potential applications in temperature management, chemical-composition monitoring, and process reliability. Future research will focus on prospective industrial validation, integration with actual measurement-release systems, and evaluation of the framework across additional EAF furnaces, production sites, and operating regimes to investigate its transferability and effectiveness in supporting operational decisions and process optimization.




\section*{CRediT Authorship Contribution Statement}

\textbf{Ahmed Shoyeb Raihan:} Conceptualization, Data curation, Formal analysis, Investigation, Methodology, Software, Validation, Visualization, Writing – original draft. \textbf{Farzana Islam:} Conceptualization, Data curation, Formal analysis, Investigation, Methodology, Writing – original draft. \textbf{Ashish Nimbarte:} Validation, Supervision, Resources, Project administration, Writing – review and editing. \textbf{Imtiaz Ahmed:} Conceptualization, Formal analysis, Investigation, Resources, Validation,  Project administration, Supervision, Writing – review and editing.

\section*{Data Availability}

The original industrial EAF dataset is publicly available through the repository cited in Section~\ref{subsec:eaf_sources}. The processed modeling data and analysis code are available from the corresponding author upon reasonable request.

\section*{Funding}

This research was generously supported by the United States Environmental Protection Agency (EPA) under the Pollution Prevention (P2) practices grant.

\section*{Acknowledgements}

The authors would like to thank the Department of Industrial and Management Systems Engineering (IMSE), West Virginia University (WVU) for their invaluable support.

\bibliographystyle{elsarticle-num} 
\bibliography{references}

\end{document}